\pdfoutput=1
\documentclass[11pt]{article}
\usepackage{acl}
\usepackage{times}
\usepackage{latexsym}
\usepackage[T1]{fontenc}
\usepackage[utf8]{inputenc}
\usepackage{microtype}
\usepackage{inconsolata}
\usepackage{lipsum}
\usepackage{stfloats}
\usepackage{graphicx}
\usepackage{paralist}
\usepackage{amsmath}
\usepackage{multirow}
\usepackage{caption}
\usepackage{subcaption}
\usepackage{xspace}
\usepackage{makecell}
\usepackage{algorithm}  
\usepackage{algorithmic}
\usepackage{listings}
\usepackage{url}
\usepackage{booktabs}
\usepackage{colortbl}
\usepackage{bbm}
\usepackage{amssymb}
\usepackage{float}
\usepackage[misc]{ifsym}

\makeatletter
\newcommand{\printfnsymbol}[1]{
  \textsuperscript{\@fnsymbol{#1}}
}
\makeatother
\usepackage[noabbrev,capitalize]{cleveref}
\newcommand{\MixChunk}{\textsc{Mix}Chunk\xspace}
\newcommand{\UniChunk}{\textsc{Uni}Chunk\xspace}
\newcommand{\BMChunk}{\textsc{Bm25}Chunk\xspace}

\newcommand{\UniChunkMask}{\textsc{Intra}Doc\xspace}

\newcommand{\intraMask}{intra-document causal masking\xspace}

\newcommand{\eg}{\textit{e.g.}\xspace}

\usepackage[symbol*]{footmisc}
\usepackage{enumitem}

\title{Analysing The Impact of Sequence Composition \\ on Language Model Pre-Training}

\author{
  Yu Zhao\printfnsymbol{2}\textsuperscript{\Letter}\quad
  Yuanbin Qu\printfnsymbol{3}\quad
  Konrad Staniszewski\printfnsymbol{4}\quad
  Szymon Tworkowski\printfnsymbol{4}\quad \\
  \bf
  Wei Liu\printfnsymbol{3}\quad 
  Piotr Miłoś\printfnsymbol{4}\quad
  Yuxiang Wu\printfnsymbol{5}\quad
  Pasquale Minervini\printfnsymbol{2}\textsuperscript{\Letter} \\
  \printfnsymbol{2}University of Edinburgh\quad
  \printfnsymbol{3}Xiaomi AI Lab\quad
  \printfnsymbol{4}University of Warsaw  \quad
  \printfnsymbol{5}Weco AI\\
  {\normalsize \tt y.zhao-203@sms.ed.ac.uk \quad  p.minervini@ed.ac.uk}
}

\begin{document}
\maketitle 
\begin{abstract}
Most language model pre-training frameworks concatenate multiple documents into fixed-length sequences and use \emph{causal masking} to compute the likelihood of each token given its context; this strategy is widely adopted due to its simplicity and efficiency.
However, to this day, the influence of the pre-training sequence composition strategy on the generalisation properties of the model remains under-explored.
In this work, we find that applying causal masking can lead to the inclusion of distracting information from previous documents during pre-training, which negatively impacts the performance of the models on language modelling and downstream tasks.
In \emph{intra-document causal masking}, the likelihood of each token is only conditioned on the previous tokens in the same document, eliminating potential distracting information from previous documents and significantly improving performance.
Furthermore, we find that concatenating related documents can reduce some potential distractions during pre-training, and our proposed efficient retrieval-based sequence construction method, \textsc{Bm25}Chunk, can improve in-context learning (+11.6\%), knowledge memorisation (+9.8\%), and context utilisation (+7.2\%) abilities of language models without sacrificing efficiency.
{\footnotetext{\scriptsize\url{https://github.com/yuzhaouoe/pretraining-data-packing}}}
\end{abstract}

\pagestyle{plain}
\setcounter{page}{1}
\pagenumbering{arabic}

\section{Introduction}

Large Language Models (LLMs) are pre-trained on large amounts of documents by optimising a language modelling objective and show an intriguing ability to solve a variety of downstream NLP tasks~\citep{gpt3,pythia,llama,mistral}.
Previous works emphasise the importance of pre-training data quality~\citep[e.g.,][]{textbooks,DBLP:conf/acl/LeeINZECC22,d4,slimpajama} and diversity~\citep[e.g.,][]{doremi,pile,minipile} to improve the generalisation properties of language models.
However, the influence of the pre-training sequence composition strategy remains largely under-explored.

\begin{figure}[t]
\centering
\label{fig:intro_packing_and_masking}
\end{figure}

For most decoder-only language model pre-training pipelines~\cite[e.g.,][]{megatron,fairseq,gpt3,pythia,easylm,llm360,tinyllama}, constructing a pre-training instance involves \emph{packing}, which refers to the process of combining randomly sampled documents into a \emph{chunk} that matches the size of the context window; and \emph{causal masking}, which refers to predicting the next token conditioned on all previous tokens, including those from different documents in the chunk.
An alternative to causal masking is \emph{\intraMask}, where the likelihood of each token is conditioned on the previous tokens from the same document; \intraMask is not commonly used in existing open-source pre-training frameworks as it is argued to adversely impact pre-training efficiency~\citep{gpt3, sparse-flash-attention}.
However, to the best of our knowledge, there is no systematic analysis in the literature on how causal masking affects the generalisation properties of models despite its role in improving efficiency.

To analyse the impact of the packing and masking strategies on pre-training, we pre-train language models using \intraMask (referred to as \UniChunkMask, \cref{sec:masking}) and compare them with models pre-trained via causal masking with several \emph{packing} strategies by varying the relatedness of the documents in the pre-training chunks.
Specifically, we analyse the results produced by a commonly used baseline method that randomly samples and packs documents (\MixChunk); a method that samples and packs documents from the same source based on their meta-information (\UniChunk); and our proposed efficient retrieval-based packing method, which 
retrieves and packs related documents (\BMChunk, \cref{sec:bm25chunk}).

Our experimental results indicate that using causal masking without considering the boundaries of documents can lead to the inclusion of distracting information from previous documents during pre-training (\cref{sec:pretraining} and \cref{sec:attention_distraction}), negatively impacting the performance of the models in downstream tasks (\cref{sec:downstream_tasks}).
We observe that \intraMask, which eliminates the potential distractions from irrelevant documents during pre-training, can significantly improve the performance of the model while increasing its runtime ($+4\%$ in our implementation, see \cref{sec:speed}).

We also find that improving the relatedness of the documents in pre-training chunks can reduce some potential distractions from previous documents (e.g., \UniChunk avoids packing documents from different distributions, such as code and news text), which can improve the performance of causal masking models on a wide array of downstream tasks.
Finally, we show that our proposed efficient retrieval-based packing method, \BMChunk, can improve a model's language modelling (+$6.8\%$), in-context learning (+$11.6\%$), knowledge memorisation (+$9.8\%$), and context utilisation (+$7.2\%$) abilities using causal masking and thus without sacrificing pre-training efficiency.

Our main contributions and findings can be summarised as follows:
\begin{itemize}[leftmargin=*,noitemsep,nolistsep]
\item We systematically analyse and compare the models pre-trained using causal masking and \intraMask; our experimental results reveal that using causal masking without considering the boundaries of documents can result in significant performance degradation (\cref{sec:pretraining} and \cref{sec:downstream_tasks}).
\item We find that improving the relatedness of the documents in each pre-training chunk benefits causal masking models, and our proposed efficient retrieval-based packing method (\BMChunk, \cref{sec:bm25chunk}) can improve the performance of language models significantly.
\item We quantitatively analyse the attention distribution of the models during language modelling (\cref{sec:attention_distraction}), and investigate the burstiness property of pre-training chunks (\cref{sec:data_distribution}); our findings indicate that models can be more robust to irrelevant contexts and obtain better performance when improving the relatedness of documents in pre-training chunks.
\end{itemize}

\section{Packing and Masking Strategies for Pre-Training Sequence Composition}
\label{sec:method}
In this section, we formally introduce the pre-training data packing strategies, causal masking, and \intraMask.

\subsection{Packing Strategies}
Let $\mathcal{D}_{i}$ represent a corpus, such as Wikipedia, C4, or GitHub, and let $\mathcal{D} = \bigcup_{s} \mathcal{D}_s$ denote the dataset resulting from the union of such corpora.
Furthermore, each corpus $\mathcal{D}_s$ is defined as a set of documents $\mathcal{D}_s = \{ d_{1}, \ldots, d_{|\mathcal{D}_s|}\}$
, where each document $d_{i}$ is defined as a sequence of tokens $d_{i} = \left( x_{1}, \ldots, x_{|d_{i}|} \right)$.

A \emph{packing strategy} involves first selecting a set of documents $\{d_i\}_{i=1}^n$ from $\mathcal{D}$, and then packing them into a chunk $C$ with a fixed length $|C|=L$.
Following \citet{gpt3}, we concatenate the documents $\{d_i\}_{i=1}^n$ by interleaving them with end-of-sentence ($\textsc{[eos]}$) tokens to construct a chunk.
A \emph{packed sequence} (or \emph{chunk}) $C$ is denoted as:
\begin{equation}
\label{eq:chunk}
C = (d_1 \textsc{[eos]} d_2 \textsc{[eos]} \ldots \textsc{Split}(d_n)),
\end{equation}
\noindent where $[\textsc{eos}]$ is the end-of-sentence token, $\textsc{Split}()$ truncates the last document such that $|C|=L$, and the content of the chunk $C$ will be removed from the dataset $\mathcal{D}$ to avoid sampling the same documents multiple times.

In the following, we introduce three strategies to sample the documents $\{d_i\}_{i=1}^n$ from the dataset $\mathcal{D}$ for composing each pre-training chunk, namely \MixChunk, \UniChunk, and \BMChunk.

\begin{figure}[t]
\small
    \centering
    \begin{subfigure}[b]{1\columnwidth}
    \includegraphics[width=1\linewidth]{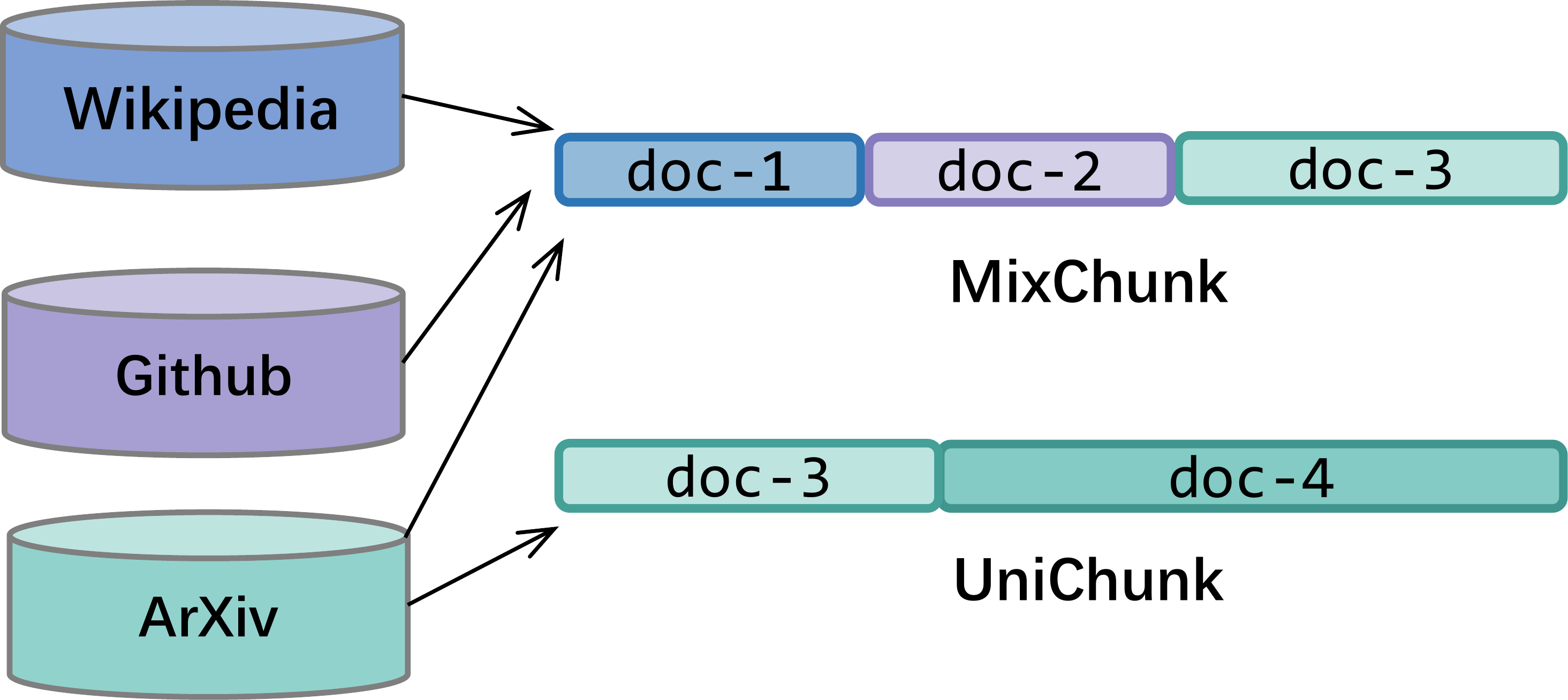}
    \caption{
    \MixChunk randomly samples documents from all corpora to construct pre-training sequences, which can pack documents from different sources. \UniChunk randomly samples documents from a single source to construct a sequence.
    }
    \label{fig:sub_a_mix_and_uni}
    \end{subfigure}
    \vspace{2mm}
\\
    \begin{subfigure}[b]{1\columnwidth}
    \includegraphics[width=1\linewidth]{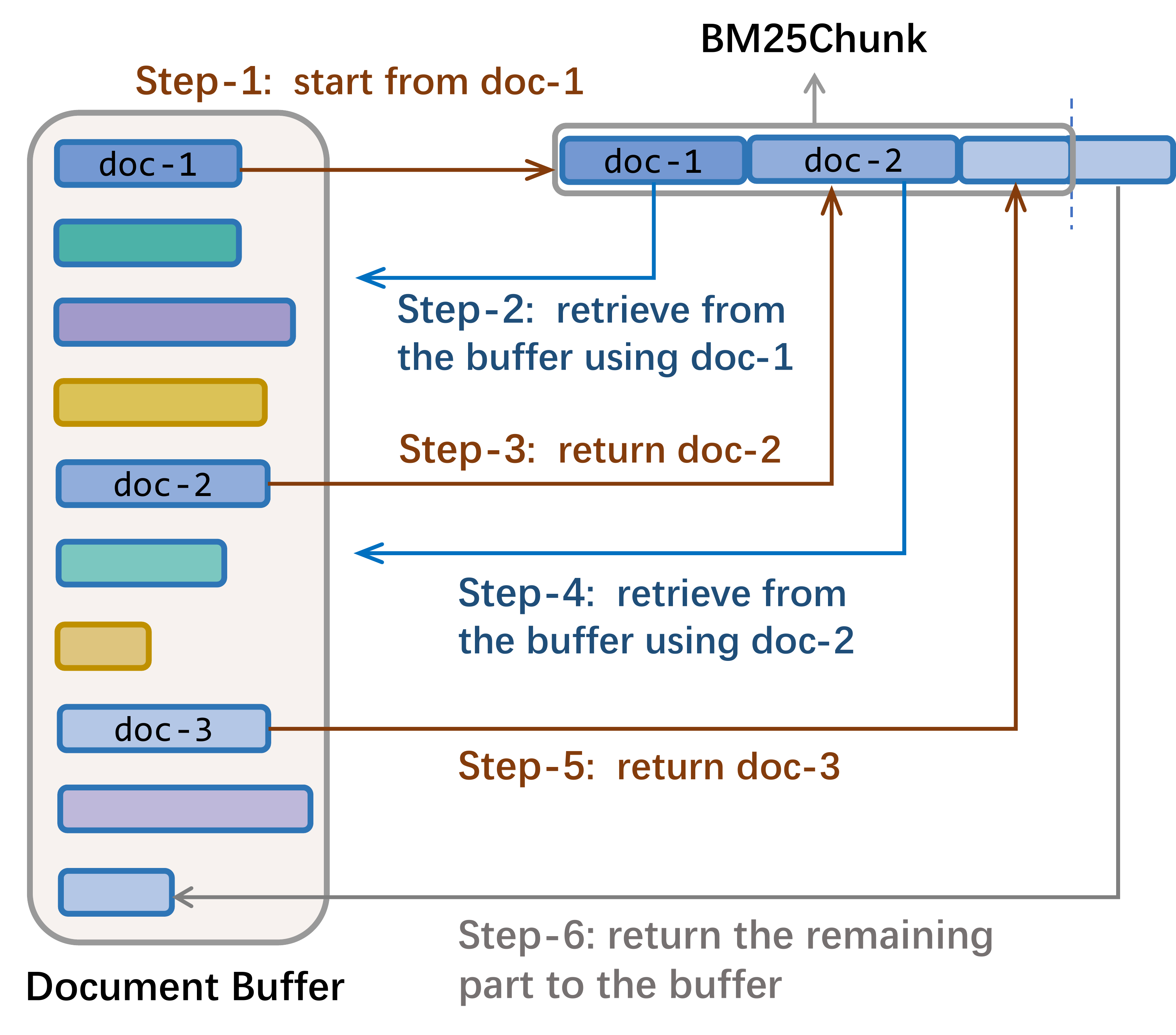}
    \caption{The sequence construction process in \BMChunk. The left part represents a document buffer that caches a set of documents randomly sampled from the corpus.}
    \label{fig:sub_b_bm25}
    \end{subfigure}
\caption{Packing strategies for pre-training chunks construction.
(a) illustrates the compositions of \MixChunk and \UniChunk; (b) presents the sequence construction process of \BMChunk.
}
    \label{fig:packing}
\end{figure}

\paragraph{\MixChunk}
In \MixChunk (baseline), documents $d_i \in \mathcal{D}$ are sampled uniformly at random from the entire pre-training corpus $\mathcal{D}$:
\begin{equation*}
d_i \sim \text{Uniform}(\mathcal{D}).
\end{equation*}
As a result, in \MixChunk, a chunk can contain documents from different source datasets, \eg, Wikipedia and GitHub, as shown in \cref*{fig:packing}(\subref{fig:sub_a_mix_and_uni}).

\paragraph{\UniChunk}
In \UniChunk, each chunk is composed of documents from a single source corpus $\mathcal{D}_s$:
\begin{equation*}
    d_i \sim \text{Uniform}(\mathcal{D}_s),\quad \text{with } \mathcal{D}_s \subseteq \mathcal{D}.
\end{equation*}
This helps to avoid packing documents from different distributions (such as code and news text) together. 
To construct a training batch, we sample sequences from each corpus $\mathcal{D}_s$ in proportion to the number of tokens in $\mathcal{D}_s$.

\paragraph{\BMChunk}
\label{sec:bm25chunk}

To improve the relevance of documents in pre-training chunks, we employ a BM25-based retriever to construct pre-training chunks, referred to as \BMChunk.
Specifically, given a document $d_i \in \mathcal{D}_s$, we retrieve a sequence of documents $\{d_i\}_{i=1}^n$ by $d_{i+1} = \textsc{Retrieve}(d_i, \mathcal{D}_s)$; here, $\textsc{Retrieve}( d_i, \mathcal{D}_s)$ retrieves the most similar documents to $d_i$ from $\mathcal{D}_s$ based on BM25 scoring.

However, this retrieval process can be computationally inefficient due to the size of the pre-training corpus $\mathcal{D}_{s}$.
To improve the efficiency of the retrieval step, we restrict the retrieval scope to a subset $\mathcal{B}_s \subseteq \mathcal{D}_s$ of the corpus $\mathcal{D}_s$, reducing the computational complexity of retrieval; the proposed approach is outlined in \cref{fig:packing}(\subref{fig:sub_b_bm25}).
More formally, we introduce a document buffer $\mathcal{B}_s \subseteq \mathcal{D}_s$ that contains $k$ documents uniformly sampled from $\mathcal{D}_s$, which serves as the retrieval source for constructing pre-training chunks:
\begin{equation*} \label{eq:bm25r}
    d_1 \sim \text{Uniform}(\mathcal{B}_s), \quad d_{i+1} = \textsc{Retrieve}( d_i, \mathcal{B}_s).
\end{equation*}
After retrieving a sequence of documents $\{d_i\}_{i=1}^n$ from the buffer $\mathcal{B}_s$ for constructing a chunk, we refill the buffer by sampling new documents from $\mathcal{D}_s$.
The time complexity analysis and more details are presented in \cref{sec:bm25-detail}.

\subsection{Masking Strategies}
\label{sec:masking}
Another core element of LLM pre-training is the \emph{masking} strategy, which determines how next-token prediction distributions are conditioned on other tokens in the sequence.
\paragraph{Causal Masking} 
In causal masking, each token in a sequence is predicted solely based on all preceding tokens in the sequence.
More formally, given a chunk $C = (x_{1}, \ldots, x_{|C|})$ defined as in \cref{eq:chunk}, the likelihood of $C$ is given by:
\begin{equation*}
P(C) = \prod_{i=1}^{|C|}  P(x_{i} \mid x_{1}, \ldots, x_{i - 1}),
\end{equation*}
\noindent where $P(x_{i} \mid x_{1}, \ldots, x_{i - 1})$ denotes the probability of the token $x_{i}$ given all preceding tokens $x_{1}, \ldots, x_{i - 1}$ in the chunk.
During pre-training, \emph{causal masking} implies that, given a chunk $C$, the probability of each token in $C$ will be conditioned on all preceding tokens, including those belonging to different documents.
Causal masking is the \emph{standard practice} when pre-training decoder-only LLMs~\cite[e.g.,][]{megatron,gpt3,opt,pythia,easylm,llm360,tinyllama}.

\paragraph{Intra-Document Causal Masking}
In \intraMask, on the other hand, the probability of each token is conditioned on the previous tokens within the same document.
More formally, given a chunk $C$ defined as in \cref{eq:chunk}, the probability of each token $d_{ij}$ belonging to document $d_{i}$ is only conditioned on the preceding tokens within $d_{i}$:
\begin{equation*}
P(C) = \prod_{i=1}^{n} \prod_{j}^{|d_{i}|} P\left(d_{ij} \mid d_{i1}, \ldots, d_{i(j-1)}\right).
\end{equation*}
We refer to models trained using \intraMask as \UniChunkMask.
The details of implementation are available in \cref{sec:speed}.

\section{Language Model Pre-Training}
\label{sec:pretraining}
\begin{table*}[t]
\centering
\small
\resizebox{\textwidth}{!}{
\begin{tabular}{clccccccc|l}
\toprule
$L$ & \bf Model & \bf CommonCrawl & \bf C4 & \bf Wikipedia & \bf GitHub & \bf StackExchange & \bf Book & \bf ArXiv & \bf Avg. \\ \midrule
\multirow[m]{4}{*}{2K} 
& \MixChunk  & $13.284$ & $13.884$ & $6.811$ & $5.531$ & $8.051$ & $11.623$ & $5.203$ & $9.172$ \\ 
& \UniChunk & $11.805$ & $\underline{13.650}$ & $6.546$ & $5.518$ & $7.839$ & $11.353$ & $5.106$ & $8.831_{\downarrow 0.341}$ \\
& \BMChunk & $\textbf{11.418}$ & $13.677$ & $\underline{6.237}$ & $\underline{4.585}$ & $\underline{7.623}$ & $\underline{11.253}$ & $\underline{5.059}$ & $\underline{8.550}_{\downarrow 0.622}$  \\ 
& \UniChunkMask & $\underline{11.631}$ & $\textbf{13.197}$ & $\textbf{6.084}$ & $\textbf{4.252}$ & $\textbf{7.535}$ & $\textbf{11.130}$ & $\textbf{5.030}$ & $\textbf{8.410}_{\downarrow 0.883}$ \\ 
\midrule
\multirow[m]{4}{*}{8K} & \MixChunk  & $9.645$ & $14.424$ & $7.010$ & $7.496$ & $8.634$ & $11.337$ & $4.911$ & $9.065$ \\ 
& \UniChunk & $9.478$ & $14.190$ & $6.897$ & $7.006$ & $8.456$ & $11.117$ & $4.812$ & $8.851_{\downarrow 0.214}$ \\ 
& \BMChunk & $\underline{9.144}$ & $\underline{13.579}$ & $\underline{6.287}$ & $\underline{5.463}$ & $\underline{8.022}$ & $\underline{10.810}$ & $\underline{4.715}$ & $\underline{8.289}_{\downarrow 0.776}$ \\ 
& \UniChunkMask & $\textbf{8.994}$ & $\textbf{13.173}$ & $\textbf{6.073}$ & $\textbf{5.010}$ & $\textbf{7.894}$ & $\textbf{10.701}$ & $\textbf{4.705}$ & $\textbf{8.079}_{\downarrow 0.986}$ \\ 
\bottomrule
\end{tabular}
}
\caption{Evaluation of perplexity on SlimPajama's test set.
The best score is highlighted in bold, and the second best is highlighted with an underline. $L$ is the maximum length of the sequence for pre-training. Subscript $_{\downarrow}$ presents the PPL improvement over the \emph{baseline} method \MixChunk. We report the results of next-token accuracy in \cref{sec:next-token-accuracy}.
}
\label{tab:pre-training-ppl}
\end{table*}
\subsection{Settings}
\paragraph{Pre-Training Corpora}
In this work, we use SlimPajama~\citep{slimpajama} as the pre-training corpus, which consists of seven sub-corpora, including CommonCrawl, C4, Wikipedia, GitHub, StackExchange, ArXiv, and Book.
This allows us to investigate packing strategies in a mixed corpora setting.
We sample documents with $150$B tokens from SlimPajama as the pre-training corpus and ensure each subset maintains the same proportion of tokens as in the original dataset.

\paragraph{Pre-Training Models}
The model implementation is based on the LLaMA~\citep{llama} architecture with minor modifications to support \intraMask.
We pre-train 1.3B parameters models using context windows of 2,048 (referred to as 2K) and 8,192 (8K) tokens.
We use the same set of documents with the difference in pre-training sequence composition to pre-train models, including causal masking models, i.e., \MixChunk, \UniChunk, and \BMChunk, and \intraMask models \UniChunkMask.
More details are available in \cref{sec:pre-training detail}.

Previous works \citep{gpt3, sparse-flash-attention} argued that dynamic sequence-specific sparse masking reduces training efficiency.
Compared to causal masking, we observe a $4.0\%$ efficiency degradation on \intraMask in our implementation, and the discussion on implementation is presented in \cref{sec:speed}.

\subsection{Results}
For evaluating LLMs trained under different packing strategies, in this work, we compute the perplexity (PPL) of a held-out set of documents where each document is treated independently.
The results are summarised in~\cref{tab:pre-training-ppl}.

We can see that \BMChunk achieves the lowest PPL among the three causal masking models, yielding a lower average PPL compared to \MixChunk in the 2K ($- 0.62$) and 8K ($- 0.78$) settings.
Furthermore, \UniChunk also yields a lower average PPL than the baseline \MixChunk ($- 0.34$ and $- 0.21$).
These results indicate that increasing the relatedness of documents in a sequence can improve the language modelling ability of models.

When considering models trained via \intraMask, we can see that \UniChunkMask achieves the lowest PPL compared to all models trained via causal masking. 
This indicates eliminating the potential distracting information from irrelevant documents during pre-training benefits the language modelling ability of models.
Specifically, we observe that both \BMChunk and \UniChunkMask obtain significantly lower PPLs on GitHub, where \UniChunkMask improves over \UniChunk in both the 2K ($- 1.3$ PPL) and 8K ($- 2.0$) models.
For \UniChunk, though we avoided packing web text and code, its improvement over \MixChunk on GitHub is slight.
This phenomenon could imply that \emph{code pre-training is more adversely affected by the distraction of unrelated context}, and both \intraMask and retrieval-based sequence construction strategy can alleviate this issue.

\section{Experiments on Downstream Tasks}
\label{sec:downstream_tasks}
In the following, we evaluate the in-context learning, knowledge memorisation, and context utilisation abilities of the models.

\subsection{In-Context Learning} \label{sec:in_context_learning}
Following \citet{in-context-pretraining}, we evaluate in-context learning abilities of the models using seven text classification datasets, namely SST2~\citep{sst2}, Amazon~\citep{yelp_amazon_agnews},  Yelp~\citep{yelp_amazon_agnews}, DBpedia~\citep{dbpedia}, AGNews~\citep{yelp_amazon_agnews}, and TweetEval hate/offensive tweet detection tasks~\citep{tweeter-eval}.

In~\cref{tab:nlu}, we report the in-context learning accuracy values of the models in few-shots learning settings, using $20$ and $48$ demonstrations for 2K and 8K models, respectively.
We truncate the input sequences to fit within their respective context windows.
For models pre-trained using causal masking, we can see that \UniChunk produces more accurate results than \MixChunk, while \BMChunk yields a higher average accuracy than \MixChunk for 2K ($+ 11.6\%$) and 8K ($+ 11.3\%$) models.
These results indicate that \emph{increasing relatedness of the documents in pre-training chunks can improve the in-context learning abilities of the models}.

\begin{table*}[ht]
\centering
\small
\setlength\tabcolsep{3.5pt}
\resizebox{\textwidth}{!}{
\begin{tabular}{cllllllll|l}
\toprule
$L$ & \bf Model  & \bf SST2 & \bf Amazon &\bf  DBpedia &\bf  AGNews & \bf Yelp &\bf  Hate & \bf Offensive & \bf Avg.\\ \midrule
\multirow[m]{4}{*}{2K} & \MixChunk & $71.53_{\pm 13.8}$ & $81.57_{\pm 15.7}$ & $40.87_{\pm 3.34}$ & $\underline{74.98}_{\pm 0.99}$ & $86.89_{\pm 4.81}$ & $47.10_{\pm 7.51}$ & $41.82_{\pm 20.46}$ & $63.54$ \\
& \UniChunk & $\underline{77.61}_{\pm 10.05}$ & $\underline{90.88}_{\pm 1.13}$ & $36.61_{\pm 2.15}$ & $70.39_{\pm 2.23}$ & $91.16_{\pm 0.35}$ & $46.20_{\pm 5.67}$ & $42.30_{\pm 14.92}$ & $65.02$ \\ 
& \BMChunk & $\textbf{83.73}_{\pm 8.17}$ & $\textbf{90.90}_{\pm 3.20}$ & $\textbf{50.16}_{\pm 2.61}$ & $\textbf{75.98}_{\pm 2.73}$ & $\underline{91.67}_{\pm 3.68}$ & $\underline{48.58}_{\pm 5.26}$ & $\underline{55.36}_{\pm 15.10}$ & $\textbf{70.91}$ \\
& \UniChunkMask & $73.65_{\pm 13.61}$ & $84.06_{\pm 12.68}$ & $\underline{46.82}_{\pm 1.82}$ & $72.32_{\pm 2.66}$ & $\textbf{91.91}_{\pm 0.97}$ & $\textbf{55.72}_{\pm 3.47}$ & $\textbf{69.14}_{\pm 5.37}$ & $\underline{70.52}$ \\
\midrule
\multirow[m]{4}{*}{8K} 
 & \MixChunk & $76.01_{\pm 8.14}$ & $87.32_{\pm 3.08}$ & $45.94_{\pm 3.70}$ & $68.21_{\pm 6.21}$ & $79.06_{\pm 9.99}$ & $42.85_{\pm 1.19}$ & $37.03_{\pm 14.28}$ & $62.43$ \\
& \UniChunk & $\textbf{81.61}_{\pm 8.63}$ & $88.30_{\pm 2.68}$ & $52.84_{\pm 2.36}$ & $63.16_{\pm 9.25}$ & $83.45_{\pm 6.41}$ & $45.50_{\pm 3.00}$ & $46.84_{\pm 16.78}$ & $65.96$ \\ 
& \BMChunk & $\underline{80.87}_{\pm 6.16}$ & $\underline{91.39}_{\pm 1.30}$ & $\underline{56.57}_{\pm 2.33}$ & $\textbf{74.79}_{\pm 2.89}$  & $\underline{85.19}_{\pm 6.93}$ & $\textbf{49.12}_{\pm 5.17}$ &  $\underline{48.33}_{\pm 15.88}$ & $\underline{69.47}$ \\ 
& \UniChunkMask & $72.38_{\pm 3.97}$ &  $\textbf{93.25}_{\pm 0.91}$ & $\textbf{61.85}_{\pm 6.89}$ & $\underline{72.49}_{\pm 4.72}$ & $\textbf{92.83}_{\pm 1.38}$ & $\underline{46.20}_{\pm 3.26}$ & $\textbf{59.59}_{\pm 9.88}$ & $\textbf{71.23}$ \\ 
\bottomrule
\end{tabular}
}
\caption{In-context learning performance evaluated by text classification accuracy across seven datasets. Accuracy and deviation (subscript) are calculated using different sets of demonstrations sampled by $16$ random seeds.
}
\label{tab:nlu}
\end{table*}

\begin{figure}[t]
    \centering
    \includegraphics[width=\columnwidth]{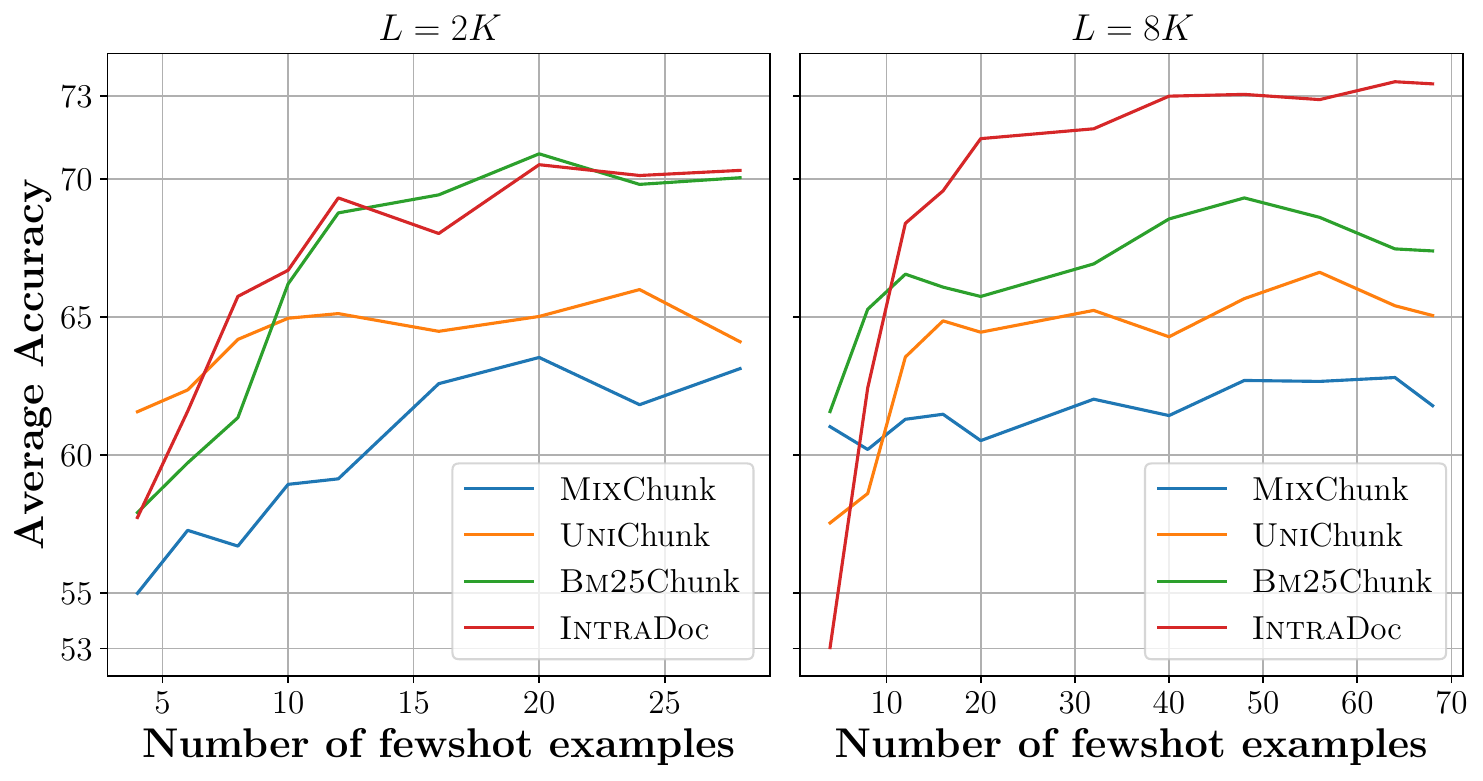}
    \caption{Average in-context learning accuracy using different numbers of few-shot demonstrations -- the left and right figures show the results of 2K and 8K models.
    }
    \label{fig:fewshot_nums}
\end{figure}

In~\cref{fig:fewshot_nums}, we present the average accuracy using different numbers of few-shot demonstrations.
We observe that \BMChunk has an on-par accuracy with \UniChunkMask on the 2K setting; however, \UniChunkMask obtains a significantly higher accuracy compared to \BMChunk on the 8K setting.
It may imply that using a longer context window size can result in increased distractions for causal masking pre-training; meanwhile, constrained by the performance of the retrieval method, \BMChunk decreases the accuracy on the 8K setting.
For 8K models, \MixChunk and \UniChunk obtain similar results to their corresponding 2K models, and they do not improve the accuracy when increasing the number of demonstrations.
It might be due to the similar levels of distraction in both 2K and 8K settings using random packing strategies.

\subsection{Knowledge Memorisation}
\label{sec:knowledge_memorisation}

\begin{table}[t]
\centering
\small
\begin{tabular}{clcc|r}
\toprule
$L$ & \bf  Model  & \bf  NQ & \bf TQA & \bf Avg.\\ \midrule
\multirow[m]{4}{*}{2K} & MixChunk & $6.19_{\pm 0.24}$ & $14.47_{\pm 0.75}$ & $10.33$ \\
& \UniChunk & $6.70_{\pm 0.26}$ & $15.53_{\pm 0.74}$ & $11.12$\\
& \BMChunk & $\underline{7.10}_{\pm 0.27}$ & $\underline{15.57}_{\pm 0.65}$ & $\underline{11.34}$\\ 
& \UniChunkMask & $\textbf{7.17}_{\pm 0.33}$ & $\textbf{16.04}_{\pm 0.35}$ & $\textbf{11.60}$\\ 
\midrule
\multirow[m]{4}{*}{8K} & MixChunk & $5.08_{\pm 0.14}$ & $10.90_{\pm 1.34}$ & $7.99$ \\
& \UniChunk & $5.25_{\pm 0.37}$ & $10.59_{\pm 1.10}$ & $7.92$ \\
& \BMChunk & $\underline{5.37}_{\pm 0.43}$ & $\underline{11.09}_{\pm 0.67}$ & $\underline{8.23}$ \\ 
& \UniChunkMask & $\textbf{6.89}_{\pm 0.08}$ & $\textbf{15.09}_{\pm 0.79}$ & $\textbf{10.99}$\\
\bottomrule
\end{tabular}
\caption{Exact Match scores on closed-book closed-book QA tasks.}
\label{tab:odqa}
\end{table}
We use two open-domain question-answering (ODQA) datasets, namely NaturalQuestions~\citep[NQ,][]{NaturalQuestions} and TriviaQA~\citep[TQA,][]{TriviaQA}, to evaluate the knowledge memorisation properties of the models.
We use $12$ demonstrations for the 2K models and $48$ demonstrations for the 8K models.
In \cref{tab:odqa}, we show the mean Exact Match (EM) scores calculated based on $5$ different sets of demonstrations.

For models trained with causal masking, we can see that \emph{increasing the relatedness of documents in pre-training chunks can improve the knowledge memorisation ability of models}.
Compared to the baseline \MixChunk, \BMChunk obtains $+9.8\%$ and $+3.0\%$ EM improvements on 2K and 8K models, respectively.
We also note that \intraMask significantly improves the knowledge memorisation ability, especially for 8K models, where \UniChunkMask improves EM by $+12.3\%$ and $+37.5\%$ over \MixChunk for 2K and 8K models, respectively.
These results support our hypothesis that reducing the distractions deriving from concatenating multiple, potentially unrelated documents in pre-training chunks can improve the knowledge memorisation ability of the models.

\subsection{Reading Comprehension and Retrieval-Augmented Generation}
\label{sec:mrc_and_rag}

\begin{table*}[t]
\centering
\small
\resizebox{1\textwidth}{!}{
\begin{tabular}{llcccccc|l}
\toprule
$L$ & \bf Model  & \bf RACE-h & \bf RACE-m & \bf SQuAD & \bf HotpotQA & \bf NQ-open & \bf TQA-open & \bf Avg.\\ \midrule

\multirow[m]{4}{*}{2K} & \MixChunk & $32.34_{\pm 0.43}$ & $42.77_{\pm 0.69}$& $36.70_{\pm 1.79}$ & $7.32_{\pm 1.31}$ & $20.00_{\pm 0.46}$ & $42.72_{\pm 1.37}$ & $30.31$ \\
 & \UniChunk & $\underline{34.01}_{\pm 0.52}$ & $43.52_{\pm 0.44}$ & $37.33_{\pm 2.31}$ & $7.12_{\pm 1.35}$ & $21.16_{\pm 0.96}$ & $42.32_{\pm 1.10}$ & $30.91$ \\
 & \BMChunk & $33.17_{\pm 0.36}$ & $\underline{44.92}_{\pm 0.46}$ & $\underline{37.91}_{\pm 1.84}$ & $\textbf{10.30}_{\pm 0.42}$ & $\textbf{22.10}_{\pm 0.91}$ & $\textbf{46.24}_{\pm 0.63}$ & $\underline{32.42}$ \\
 & \UniChunkMask & $\textbf{34.49}_{\pm 0.56}$ & $\textbf{44.96}_{\pm 0.59}$ & $\textbf{39.91}_{\pm 1.48}$ & $\underline{8.29}_{\pm 1.27}$ & $\underline{21.66}_{\pm 0.85}$ & $\underline{45.67}_{\pm 1.02}$ & $\textbf{32.49}$ \\
\midrule
\multirow[m]{4}{*}{8K} & \MixChunk & $31.66_{\pm 0.47}$ & $41.57_{\pm 0.44}$ & $32.79_{\pm 1.56}$ & $10.53_{\pm 0.70}$ & $20.53_{\pm 0.58}$ & $40.53_{\pm 1.03}$ & $29.60$ \\
& \UniChunk & $31.68_{\pm 0.94}$ & $41.64_{\pm 0.55}$ & $34.94_{\pm 1.84}$ & $10.57_{\pm 1.13}$ & $21.76_{\pm 0.80}$ & $39.60_{\pm 1.77}$ & $30.03$   \\
& \BMChunk & $\underline{32.63}_{\pm 0.68}$ & $\underline{44.14}_{\pm 0.48}$ & $\underline{39.45}_{\pm 1.05}$ & $\textbf{14.46}_{\pm 0.93}$ & $\underline{22.17}_{\pm 1.02}$ & $\underline{43.40}_{\pm 0.38}$ & $\textbf{34.54}$ \\
& \UniChunkMask & $\textbf{33.17}_{\pm 0.37}$ & $\textbf{45.56}_{\pm 0.38}$ & $\textbf{41.32}_{\pm 2.28}$ & $\underline{12.60}_{\pm 1.49}$ & $\textbf{22.25}_{\pm 0.13}$ & $\textbf{44.19}_{\pm 0.60}$ & $\underline{33.18}$ \\ 
\bottomrule
\end{tabular}
}
\caption{Evaluation results of machine reading comprehension and retrieval-augmented generation tasks.
}
\label{tab:mrc_and_rag}
\end{table*}

\begin{figure}[t]
\centering
    \includegraphics[width=0.95\columnwidth]{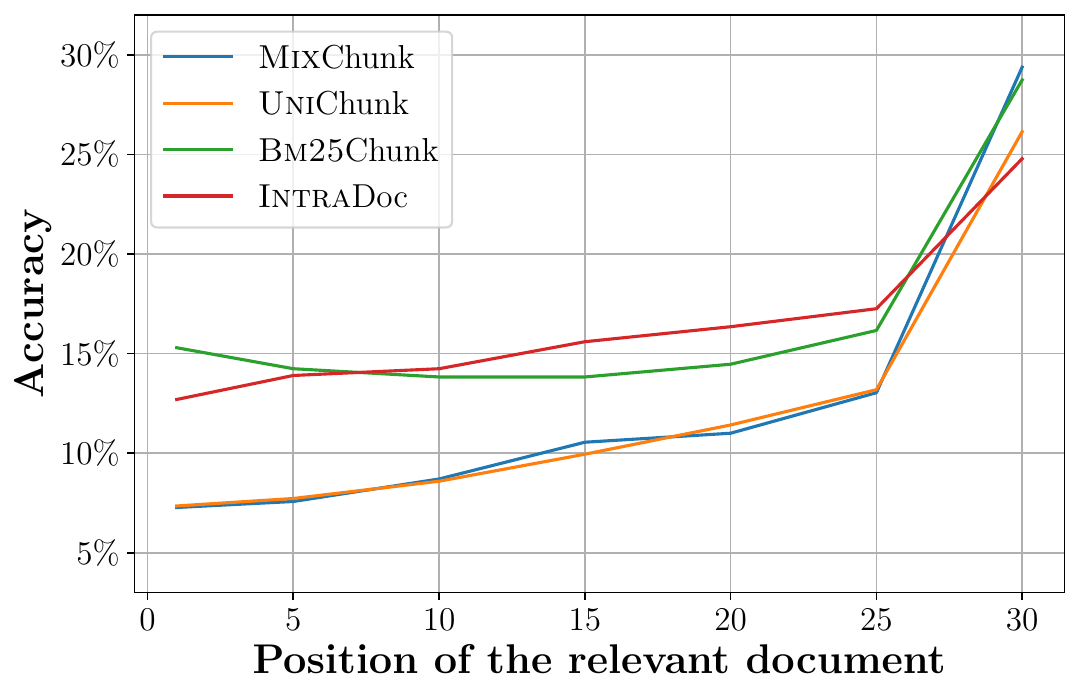}
    \caption{Accuracy on Multi-Document Question-Answering (MDQA). 
    The $x$-axis represents the position of the document that contains the answer. The $y$-axis presents the accuracy for a position.}
    \label{fig:mdqa}
\end{figure}

We evaluate the pre-trained models on a set of reading comprehension tasks, namely RACE~\citep{race}, SQuAD~\citep{squad}, HotpotQA~\citep{hotpotqa}, and the following retrieval-augmented generation (RAG) tasks: NQ, TQA, and Multi-Document Question-Answering \cite[MDQA,][]{liu2023lost}.
For NQ and TQA, we use the top two passages retrieved by the dense retriever~\citep{DPR, FiD}, denoted as NQ-open and TQA-open.
Our results for RACE, SQuAD, and RAG tasks are summarised in \cref{tab:mrc_and_rag}, while the results on MDQA are available in \cref{fig:mdqa}.

We can see that \BMChunk produces more accurate results than \MixChunk and \UniChunk in all tasks and obtains the best average accuracy, showing that \emph{increasing the relatedness of documents in pre-training chunks can improve the context utilisation ability}.
Specifically, \BMChunk obtains a significantly better accuracy on multi-hop QA task HotpotQA, showing it can better utilise multiple relevant information from the context.

\UniChunkMask obtains the best average accuracy in the 2K models and obtains the best scores in 5 of 6 tasks in the 8K models.
It indicates that eliminating potential distractions from unrelated documents and \emph{learning each document independently can improve context utilisation ability}.
This finding is different from the ideas in previous works, which suggested that pre-training with multiple documents in one context \citep{in-context-pretraining} and adding distraction in context during pre-training \citep{longllama} benefit context utilisation ability.

In MDQA, for each question, there are $30$ documents provided in the context, where only one of them contains the answer to the question --- MDQA is used to evaluate the ability of models to filter out irrelevant information and identify the relevant parts of a long context.
This task has been used to analyse the \emph{lost-in-the-middle} phenomenon in LLMs where they struggle to retrieve information stored in the middle of long contexts~\cite{liu2023lost}.
In the following, we analyse how the accuracy of models varies with the position of relevant information in the context.
In these experiments, we focus on 8K models due to their ability to handle long contexts.
The zero-shot results on MDQA  are outlined in \cref{fig:mdqa}.
We observe that both \BMChunk and \UniChunkMask tend to produce more accurate predictions than \MixChunk and \UniChunk when the relevant passage is located at the beginning or middle of the context.
These results show that \BMChunk and \UniChunkMask \emph{can better filter irrelevant context and locate relevant information}; these results are further corroborated by our experiments in \cref{sec:attention_distraction} where we analyse the attention distribution of the models during the language modelling process.

\begin{figure*}[htb]
\scriptsize
    \centering
    \begin{subfigure}[t]{0.24\textwidth}
    \includegraphics[width=1\linewidth]{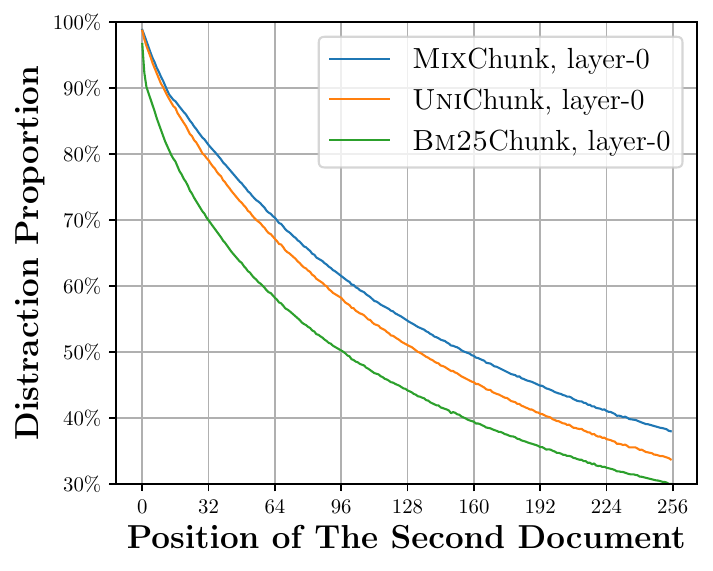}
    \captionsetup{width=0.9\linewidth}
    \caption{\scriptsize The distraction proportion of the 
    \emph{first layer}; different documents are separated by \textsc{[eos]}.}
    \label{fig:distract_a}
    \end{subfigure}
    \begin{subfigure}[t]{0.24\textwidth}
    \includegraphics[width=1\linewidth]{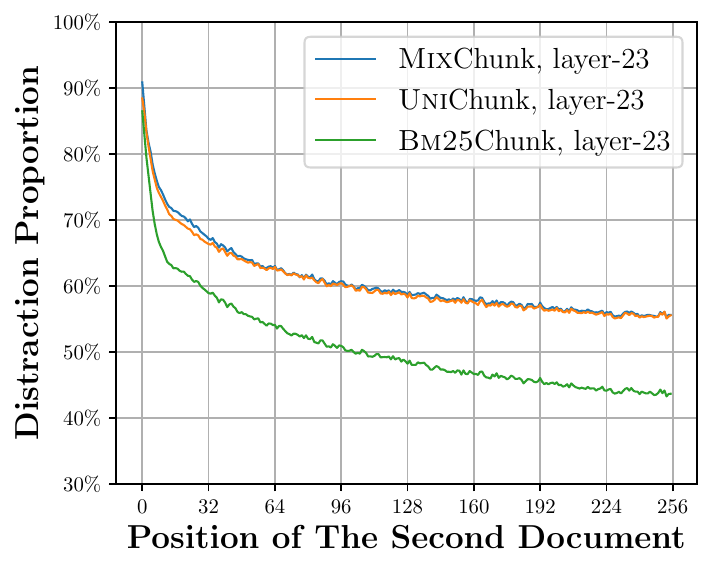}
    \captionsetup{width=0.9\linewidth}
    \caption{\scriptsize The distraction proportion of the \emph{last layer}; different documents are separated by \textsc{[eos]}.}
    \label{fig:distract_b}
    \end{subfigure}
    \begin{subfigure}[t]{0.24\textwidth}
    \includegraphics[width=1\linewidth]{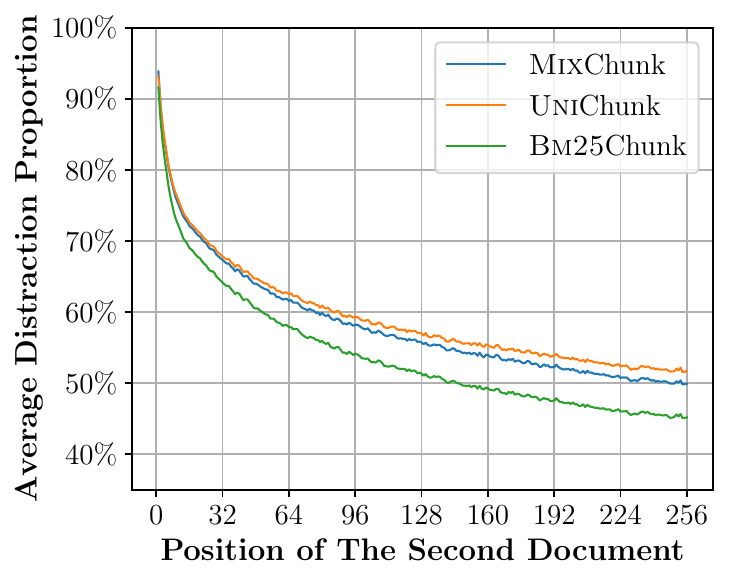}
    \captionsetup{width=0.9\linewidth}
    \caption{\scriptsize The average distraction proportion over layers; different documents are separated by \textsc{[eos]}.}
    \label{fig:distract_c}
    \end{subfigure}
    \begin{subfigure}[t]{0.24\textwidth}
    \includegraphics[width=1\linewidth]{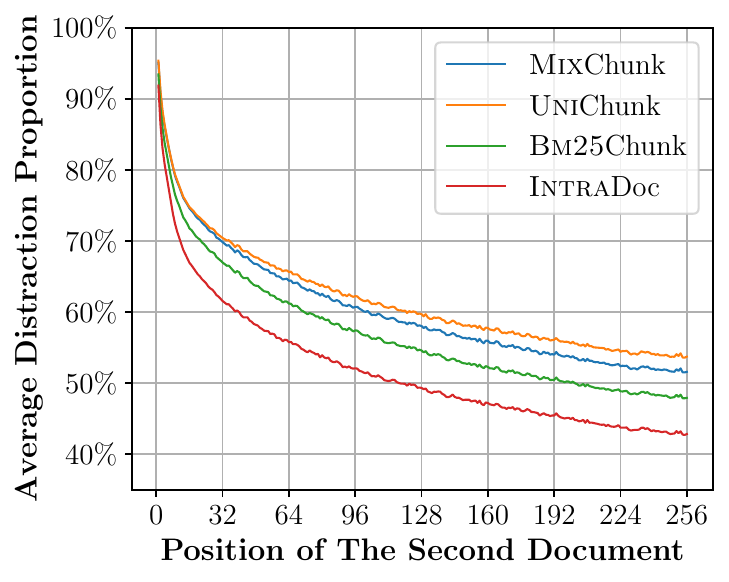}
    \captionsetup{width=0.9\linewidth}
    \caption{\scriptsize The average distraction proportion over layers; different documents are separated by "$\backslash \text{n}$"}
    \label{fig:distract_d}
    \end{subfigure}
    \caption{Distracted attention proportions of models. The $x$-axis presents the token position of the second document; the $y$-axis presents the distraction proportion calculated by \cref{eq:distraction}. Figures (a) and (b) show the distraction proportion of the first and last layers. Figures (c) and (d) are the average distraction proportion over layers. In Figure (d), we separate documents by a newline token ("$\backslash \text{n}$") and present the distraction proportion of \UniChunkMask. The results are averaged from $4096$ examples. More analysis is presented in \cref{sec:more_distraction}.}
    \label{fig:distraction}
\end{figure*}

\section{Discussion and Analysis}
\label{sec:discussion_and_analysis}

\subsection{Can Models Ignore Irrelevant Contexts Before the End-of-Sequence Token?}
\label{sec:attention_distraction}

In the following, we analyse whether models can filter irrelevant context during language modelling by examining the attention score distributions over the context.
Specifically, we concatenate two randomly sampled documents from the SlimPajama validation set, separate them by an end-of-sequence token \textsc{[eos]}, and check to which extent the attention distributions of the model focus on the irrelevant document in the sequence.
More formally, we define the \emph{distraction proportion} of the token in position $p$ in the current document at layer $l$ as:
\begin{equation} \label{eq:distraction}
    \textsc{DistrProp}(l,p) = \sum_{i=1}^{|C_{d}|} a_{p,i}^l
\end{equation}
\noindent where $|C_{d}|$ denotes the number of tokens in the irrelevant document, $a_{p,i}^l$ is the average multi-head attention scores to the $i$-th token in the irrelevant document $C_d$ at layer $l$, and $\sum_{i=1}^{|C_{d}| + p} a_{p,i}^l =1$.
In our experiments, we set $|C_d|=256$, and the results are outlined in \cref{fig:distraction}.

We can see that the latter positions have lower distraction proportions but remain $45\%$-$52\%$ average distraction proportion until the $256$th token of the second document, as shown in \cref*{fig:distraction}(\subref{fig:distract_c}).
We find that models trained via \BMChunk (green line) tend to have lower distraction proportions than other causal masking models, showing that they can better recognise relevant information in the context, matching the results in \cref{fig:mdqa}.
The above analysis also demonstrates that during the pre-training, causal masking models can be distracted by unrelated documents in context, and the models can be more robust to irrelevant contexts when reducing distractions in pre-training sequences.

Furthermore, in \cref*{fig:distraction}(\subref{fig:distract_d}), we compare \UniChunkMask and causal masking models using "$\backslash \text{n}$" as the separator instead of \textsc{[eos]}, because \textsc{[eos]} can only appear at the end of sequences during pre-training using \intraMask.
The results indicate that \UniChunkMask has the lowest distraction proportion compared to causal masking models; meanwhile, \BMChunk consistently has a lower distraction proportion than \MixChunk and \UniChunk using "$\backslash \text{n}$" as the separator.
These results match the finding in \cref{sec:mrc_and_rag}, where \UniChunkMask and \BMChunk can better recognise relevant information in the context.

\subsection{Burstiness Property of Sequences} \label{sec:data_distribution}
\citet{data-distributional-properties, supportive-pretraining-data} found a positive correlation between the model's in-context learning ability and \emph{burstiness} property of the training sequences.
Here, burstiness refers to the phenomenon where certain types of tokens occur in clusters or bursts rather than being uniformly distributed across all documents.
Burstiness is an inherent property of text; for example, a specific medical term might be frequently used in medical articles and rarely appear in general texts.
Higher burstiness results in a lower Zipf's coefficient of token frequency \emph{within a sequence}~\citep{supportive-pretraining-data}. 

Following \citet{supportive-pretraining-data}, we use Zipf's coefficient to measure the burstiness property of pre-training sequences.
Formally, let $r$ denote the rank of a token in a sequence, and $f$ is a frequency function that maps the rank $r$ to the frequency of that token in the sequence.
Then, according to Zipf's law, we have that $f(r;\alpha) \propto \frac{1}{r^\alpha}$,
where $\alpha \in \mathbb{R}^+$ is the Zipf's coefficient; a lower $\alpha$ presents an increased burstiness property within the sequence.

In \cref{tab:distribution}, we show the Zipf's coefficients $\alpha$ on different pre-training sequences.
Our results show that, for causal masking approaches that use the same chunk size, a lower Zipf's coefficient, which denotes increased burstiness property, often results in more accurate results.
However, \UniChunkMask can obtain significantly better results than \UniChunk with the same Zipf's coefficient.
The above results indicate that, for causal masking approaches, \emph{the correlation between higher burstiness and better performance could derive from reduced distractions in pre-training chunks}.
We report additional evidence for the burstiness property in \cref{sec:appendix-distribution-property}.

Note that duplication in pre-training sequences can also result in increased burstiness property, which may negatively impact the performance of language models. 
We analyse the distinct n-gram phrases of pre-training sequences in \cref{sec:appendix-distribution-property} and will investigate the impact of duplication using different pre-training corpora in future work.
\begin{table}[t]
\centering
\small
\resizebox{\linewidth}{!}{
\begin{tabular}{clccc}
\toprule
$L$ &\bf Method  & \makecell{\textbf{Zipf's} \\ \textbf{ Coeffeicient} \\ ($\alpha$)} & \makecell{\textbf{In-Context} \\ \textbf{Learning} \\ (Acc.)} & \makecell{\textbf{Knowledge} \\ \textbf{Memorisation} \\ (EM)}   \\ \midrule
\multirow[m]{3}{*}{2K} & \MixChunk & $2.122$ &  $63.54$ & $10.33$\\ 
& \UniChunk & $2.119$ & $65.02$ & $11.12$ \\ 
& \BMChunk & $2.107$ & $70.91$ & $11.34$ \\  
\midrule
\multirow[m]{3}{*}{8K} & \MixChunk & $1.976$ &  $62.43$ & $7.99$ \\ 
& \UniChunk & $1.951$ & $65.96$ & $7.92$ \\ 
& \BMChunk & $1.925$ & $69.47$ & $8.23$\\  
\midrule
2K & \UniChunkMask & $2.119$ & $70.52$ & $11.60$ \\
8K & \UniChunkMask & $1.952$ & $71.23$ & $10.99$\\  
\bottomrule
\end{tabular}
}
\caption{Zipf's coefficients of token frequency in different data. In-context learning and knowledge memorisation abilities are evaluated in \cref{sec:downstream_tasks}.
}
\label{tab:distribution}
\vspace{-5pt}
\end{table}

\section{Related Works}
\paragraph{Instance-Level Pre-training Data Composition}
GPT-3~\citep{gpt3} was pre-trained by packed documents with causal masking, with the idea that not adopting any dynamic masking can improve pre-training efficiency.
Current open-source pre-training frameworks, such as MegatronLM~\citep{megatron}, \textsc{fairseq}~\citep{fairseq}, EasyLM~\citep{easylm}, LLM360~\citep{llm360}, also follow this strategy for pre-training.
In \citet{pretraining-example-design}, authors pair similar sentences within the same sequence, while \citet{pre-train-to-learn-in-context} propose packing documents that contain similar intrinsic tasks for continual pre-training, improving the in-context learning ability of models.
Recently, \citet{in-context-pretraining} emphasise that packing relevant documents can enhance language models' in-context learning and context utilisation; however, our findings indicate that packing documents can adversely affect performance, and learning each document independently using \intraMask can reduce the distraction and improve the performance.

\paragraph{Distribution Properties of Pre-Training Data}
\citet{data-distributional-properties}~shows several data distribution properties can drive in-context learning ability, e.g., large numbers of long-tail classes, dynamic meanings of inputs, and Zipf's distribution of class frequency.
\citet{supportive-pretraining-data} used a gradient-guided method to select small-scale data for continual pre-training, showing data exhibiting burstiness properties can enhance in-context learning performance.

\paragraph{Pre-training Data Quality}
\citet{textbooks} selected high-quality data to pre-train a small-size coding model, achieving comparable performance with larger models.
\citet{DBLP:conf/naacl/ShinLAKKKCLPHS22, pile} emphasised the importance of pre-training data diversity.
\citet{DBLP:conf/acl/LeeINZECC22, d4, slimpajama, semdedup} showed the importance of data de-duplication on models' generalisation.
In our work, we use a diverse and high-quality pre-training dataset, namely SlimPajama~\citep{slimpajama}, to highlight the importance of the sequence composition strategy on language model pre-training.

\section{Conclusion}
In this work, we investigate the impact of pre-training sequence composition by pre-training models from scratch.
First, we find causal masking can result in unrelated documents distracting language modelling pre-training and hurting the performance on downstream tasks; we show that \intraMask can significantly improve the performance while decreasing the pre-training efficiency.
Second, we find improving the relatedness of documents in pre-training chunks for causal masking pre-training can reduce some potential distractions in chunks; our proposed efficient retrieval-based packing method \BMChunk can significantly improve the performance of language models without reducing pre-training efficiency.

\section*{Limitations}
\paragraph{Efficiency of Intra-Document Causal Masking} 
We show that \intraMask is an effective method to improve the performance while decreasing the pre-training efficiency.
We use FlashAttention2~\citep{flash-attention} to implement \intraMask masking without sacrificing too much efficiency (discussed in \cref{sec:speed}). Still, we do not propose a method to solve this efficiency issue completely.

\paragraph{Objective of Sequences Construction.} We discuss sequence construction methods, showing the importance of sequence compositions on the performance of models, but these methods lack an objective during sequence construction.
Since specific data distribution properties may be related to models' performance, we will explore using indicators of distributional properties to guide sequence construction in future works.

\paragraph{Scaling The Size of Language Models.}
Limited by the computation resources, we cannot conduct experiments on larger models with more pre-training steps, and different results might be drawn when increasing the models at a specific scale.
However, this work could be directly valuable for investigating pre-training relatively small models that aim at facilitating the use of language models under resource-constrained conditions.

\section*{Acknowledgements}
PM was partially funded by 
ELIAI (The Edinburgh Laboratory for Integrated Artificial Intelligence), EPSRC (grant no. EP/W002876/1), an industry grant from Cisco, and a donation from Accenture LLP; and is grateful to NVIDIA for the GPU donations.
This work was supported by the Edinburgh International Data Facility (EIDF) and the Data-Driven Innovation Programme at the University of Edinburgh.
The authors extend their gratitude to Xiaomi AI Lab for their GPU donations and assistance; as well as to Tri Dao, Piotr Nawrot, Giwon Hong, Xiaotang Du, and Aryo Gema for their help and feedback.

\bibliography{references}

\begin{thebibliography}{43}
\expandafter\ifx\csname natexlab\endcsname\relax\def\natexlab#1{#1}\fi

\bibitem[{Abbas et~al.(2023)Abbas, Tirumala, Simig, Ganguli, and Morcos}]{semdedup}
Amro Abbas, Kushal Tirumala, Daniel Simig, Surya Ganguli, and Ari~S. Morcos. 2023.
\newblock \href {https://doi.org/10.48550/ARXIV.2303.09540} {Semdedup: Data-efficient learning at web-scale through semantic deduplication}.
\newblock \emph{CoRR}, abs/2303.09540.

\bibitem[{Barbieri et~al.(2020)Barbieri, Camacho{-}Collados, Anke, and Neves}]{tweeter-eval}
Francesco Barbieri, Jos{\'{e}} Camacho{-}Collados, Luis~Espinosa Anke, and Leonardo Neves. 2020.
\newblock \href {https://doi.org/10.18653/V1/2020.FINDINGS-EMNLP.148} {Tweeteval: Unified benchmark and comparative evaluation for tweet classification}.
\newblock In \emph{Findings of the Association for Computational Linguistics: {EMNLP} 2020, Online Event, 16-20 November 2020}, volume {EMNLP} 2020 of \emph{Findings of {ACL}}, pages 1644--1650. Association for Computational Linguistics.

\bibitem[{Biderman et~al.(2023)Biderman, Schoelkopf, Anthony, Bradley, O'Brien, Hallahan, Khan, Purohit, Prashanth, Raff, Skowron, Sutawika, and van~der Wal}]{pythia}
Stella Biderman, Hailey Schoelkopf, Quentin Anthony, Herbie Bradley, Kyle O'Brien, Eric Hallahan, Mohammad~Aflah Khan, Shivanshu Purohit, USVSN~Sai Prashanth, Edward Raff, Aviya Skowron, Lintang Sutawika, and Oskar van~der Wal. 2023.
\newblock \href {https://doi.org/10.48550/ARXIV.2304.01373} {Pythia: {A} suite for analyzing large language models across training and scaling}.
\newblock \emph{CoRR}, abs/2304.01373.

\bibitem[{Brown et~al.(2020)Brown, Mann, Ryder, Subbiah, Kaplan, Dhariwal, Neelakantan, Shyam, Sastry, Askell, Agarwal, Herbert{-}Voss, Krueger, Henighan, Child, Ramesh, Ziegler, Wu, Winter, Hesse, Chen, Sigler, Litwin, Gray, Chess, Clark, Berner, McCandlish, Radford, Sutskever, and Amodei}]{gpt3}
Tom~B. Brown, Benjamin Mann, Nick Ryder, Melanie Subbiah, Jared Kaplan, Prafulla Dhariwal, Arvind Neelakantan, Pranav Shyam, Girish Sastry, Amanda Askell, Sandhini Agarwal, Ariel Herbert{-}Voss, Gretchen Krueger, Tom Henighan, Rewon Child, Aditya Ramesh, Daniel~M. Ziegler, Jeffrey Wu, Clemens Winter, Christopher Hesse, Mark Chen, Eric Sigler, Mateusz Litwin, Scott Gray, Benjamin Chess, Jack Clark, Christopher Berner, Sam McCandlish, Alec Radford, Ilya Sutskever, and Dario Amodei. 2020.
\newblock \href {https://proceedings.neurips.cc/paper/2020/hash/1457c0d6bfcb4967418bfb8ac142f64a-Abstract.html} {Language models are few-shot learners}.
\newblock In \emph{Advances in Neural Information Processing Systems 33: Annual Conference on Neural Information Processing Systems 2020, NeurIPS 2020, December 6-12, 2020, virtual}.

\bibitem[{Chan et~al.(2022)Chan, Santoro, Lampinen, Wang, Singh, Richemond, McClelland, and Hill}]{data-distributional-properties}
Stephanie Chan, Adam Santoro, Andrew~K. Lampinen, Jane Wang, Aaditya Singh, Pierre~H. Richemond, James~L. McClelland, and Felix Hill. 2022.
\newblock \href {http://papers.nips.cc/paper\_files/paper/2022/hash/77c6ccacfd9962e2307fc64680fc5ace-Abstract-Conference.html} {Data distributional properties drive emergent in-context learning in transformers}.
\newblock In \emph{NeurIPS}.

\bibitem[{Dao(2023)}]{flash-attention}
Tri Dao. 2023.
\newblock \href {https://doi.org/10.48550/ARXIV.2307.08691} {Flashattention-2: Faster attention with better parallelism and work partitioning}.
\newblock \emph{CoRR}, abs/2307.08691.

\bibitem[{Gao et~al.(2021)Gao, Biderman, Black, Golding, Hoppe, Foster, Phang, He, Thite, Nabeshima, Presser, and Leahy}]{pile}
Leo Gao, Stella Biderman, Sid Black, Laurence Golding, Travis Hoppe, Charles Foster, Jason Phang, Horace He, Anish Thite, Noa Nabeshima, Shawn Presser, and Connor Leahy. 2021.
\newblock \href {http://arxiv.org/abs/2101.00027} {The pile: An 800gb dataset of diverse text for language modeling}.
\newblock \emph{CoRR}, abs/2101.00027.

\bibitem[{Geng(2023)}]{easylm}
Xinyang Geng. 2023.
\newblock \href {https://github.com/young-geng/EasyLM} {Easylm: A simple and scalable training framework for large language models}.

\bibitem[{Gu et~al.(2023)Gu, Dong, Wei, and Huang}]{pre-train-to-learn-in-context}
Yuxian Gu, Li~Dong, Furu Wei, and Minlie Huang. 2023.
\newblock \href {https://doi.org/10.18653/V1/2023.ACL-LONG.267} {Pre-training to learn in context}.
\newblock In \emph{Proceedings of the 61st Annual Meeting of the Association for Computational Linguistics (Volume 1: Long Papers), {ACL} 2023, Toronto, Canada, July 9-14, 2023}, pages 4849--4870. Association for Computational Linguistics.

\bibitem[{Gunasekar et~al.(2023)Gunasekar, Zhang, Aneja, Mendes, Giorno, Gopi, Javaheripi, Kauffmann, de~Rosa, Saarikivi, Salim, Shah, Behl, Wang, Bubeck, Eldan, Kalai, Lee, and Li}]{textbooks}
Suriya Gunasekar, Yi~Zhang, Jyoti Aneja, Caio C{\'{e}}sar~Teodoro Mendes, Allie~Del Giorno, Sivakanth Gopi, Mojan Javaheripi, Piero Kauffmann, Gustavo de~Rosa, Olli Saarikivi, Adil Salim, Shital Shah, Harkirat~Singh Behl, Xin Wang, S{\'{e}}bastien Bubeck, Ronen Eldan, Adam~Tauman Kalai, Yin~Tat Lee, and Yuanzhi Li. 2023.
\newblock \href {https://doi.org/10.48550/ARXIV.2306.11644} {Textbooks are all you need}.
\newblock \emph{CoRR}, abs/2306.11644.

\bibitem[{Han et~al.(2023)Han, Simig, Mihaylov, Tsvetkov, Celikyilmaz, and Wang}]{supportive-pretraining-data}
Xiaochuang Han, Daniel Simig, Todor Mihaylov, Yulia Tsvetkov, Asli Celikyilmaz, and Tianlu Wang. 2023.
\newblock \href {https://doi.org/10.18653/V1/2023.ACL-LONG.708} {Understanding in-context learning via supportive pretraining data}.
\newblock In \emph{Proceedings of the 61st Annual Meeting of the Association for Computational Linguistics (Volume 1: Long Papers), {ACL} 2023, Toronto, Canada, July 9-14, 2023}, pages 12660--12673. Association for Computational Linguistics.

\bibitem[{Hoffmann et~al.(2022)Hoffmann, Borgeaud, Mensch, Buchatskaya, Cai, Rutherford, de~Las~Casas, Hendricks, Welbl, Clark, Hennigan, Noland, Millican, van~den Driessche, Damoc, Guy, Osindero, Simonyan, Elsen, Rae, Vinyals, and Sifre}]{chinchilla}
Jordan Hoffmann, Sebastian Borgeaud, Arthur Mensch, Elena Buchatskaya, Trevor Cai, Eliza Rutherford, Diego de~Las~Casas, Lisa~Anne Hendricks, Johannes Welbl, Aidan Clark, Tom Hennigan, Eric Noland, Katie Millican, George van~den Driessche, Bogdan Damoc, Aurelia Guy, Simon Osindero, Karen Simonyan, Erich Elsen, Jack~W. Rae, Oriol Vinyals, and Laurent Sifre. 2022.
\newblock \href {https://doi.org/10.48550/ARXIV.2203.15556} {Training compute-optimal large language models}.
\newblock \emph{CoRR}, abs/2203.15556.

\bibitem[{Izacard et~al.(2022)Izacard, Caron, Hosseini, Riedel, Bojanowski, Joulin, and Grave}]{contriever}
Gautier Izacard, Mathilde Caron, Lucas Hosseini, Sebastian Riedel, Piotr Bojanowski, Armand Joulin, and Edouard Grave. 2022.
\newblock \href {https://openreview.net/forum?id=jKN1pXi7b0} {Unsupervised dense information retrieval with contrastive learning}.
\newblock \emph{Trans. Mach. Learn. Res.}, 2022.

\bibitem[{Izacard and Grave(2021)}]{FiD}
Gautier Izacard and Edouard Grave. 2021.
\newblock \href {https://doi.org/10.18653/V1/2021.EACL-MAIN.74} {Leveraging passage retrieval with generative models for open domain question answering}.
\newblock In \emph{Proceedings of the 16th Conference of the European Chapter of the Association for Computational Linguistics: Main Volume, {EACL} 2021, Online, April 19 - 23, 2021}, pages 874--880. Association for Computational Linguistics.

\bibitem[{Jiang et~al.(2023)Jiang, Sablayrolles, Mensch, Bamford, Chaplot, Casas, Bressand, Lengyel, Lample, Saulnier et~al.}]{mistral}
Albert~Q Jiang, Alexandre Sablayrolles, Arthur Mensch, Chris Bamford, Devendra~Singh Chaplot, Diego de~las Casas, Florian Bressand, Gianna Lengyel, Guillaume Lample, Lucile Saulnier, et~al. 2023.
\newblock Mistral 7b.
\newblock \emph{arXiv preprint arXiv:2310.06825}.

\bibitem[{Johnson et~al.(2019)Johnson, Douze, and J{\'e}gou}]{faiss}
Jeff Johnson, Matthijs Douze, and Herv{\'e} J{\'e}gou. 2019.
\newblock Billion-scale similarity search with {GPUs}.
\newblock \emph{IEEE Transactions on Big Data}, 7(3):535--547.

\bibitem[{Joshi et~al.(2017)Joshi, Choi, Weld, and Zettlemoyer}]{TriviaQA}
Mandar Joshi, Eunsol Choi, Daniel Weld, and Luke Zettlemoyer. 2017.
\newblock \href {https://doi.org/10.18653/v1/P17-1147} {{T}rivia{QA}: A large scale distantly supervised challenge dataset for reading comprehension}.
\newblock In \emph{Proceedings of the 55th Annual Meeting of the Association for Computational Linguistics (Volume 1: Long Papers)}, pages 1601--1611, Vancouver, Canada. Association for Computational Linguistics.

\bibitem[{Kaddour(2023)}]{minipile}
Jean Kaddour. 2023.
\newblock \href {https://doi.org/10.48550/ARXIV.2304.08442} {The minipile challenge for data-efficient language models}.
\newblock \emph{CoRR}, abs/2304.08442.

\bibitem[{Karpukhin et~al.(2020)Karpukhin, Oguz, Min, Lewis, Wu, Edunov, Chen, and Yih}]{DPR}
Vladimir Karpukhin, Barlas Oguz, Sewon Min, Patrick Lewis, Ledell Wu, Sergey Edunov, Danqi Chen, and Wen-tau Yih. 2020.
\newblock \href {https://doi.org/10.18653/v1/2020.emnlp-main.550} {Dense passage retrieval for open-domain question answering}.
\newblock In \emph{Proceedings of the 2020 Conference on Empirical Methods in Natural Language Processing (EMNLP)}, pages 6769--6781, Online. Association for Computational Linguistics.

\bibitem[{Kwiatkowski et~al.(2019)Kwiatkowski, Palomaki, Redfield, Collins, Parikh, Alberti, Epstein, Polosukhin, Devlin, Lee, Toutanova, Jones, Kelcey, Chang, Dai, Uszkoreit, Le, and Petrov}]{NaturalQuestions}
Tom Kwiatkowski, Jennimaria Palomaki, Olivia Redfield, Michael Collins, Ankur Parikh, Chris Alberti, Danielle Epstein, Illia Polosukhin, Jacob Devlin, Kenton Lee, Kristina Toutanova, Llion Jones, Matthew Kelcey, Ming-Wei Chang, Andrew~M. Dai, Jakob Uszkoreit, Quoc Le, and Slav Petrov. 2019.
\newblock \href {https://doi.org/10.1162/tacl_a_00276} {Natural questions: A benchmark for question answering research}.
\newblock \emph{Transactions of the Association for Computational Linguistics}, 7:452--466.

\bibitem[{Lai et~al.(2017)Lai, Xie, Liu, Yang, and Hovy}]{race}
Guokun Lai, Qizhe Xie, Hanxiao Liu, Yiming Yang, and Eduard~H. Hovy. 2017.
\newblock \href {https://doi.org/10.18653/V1/D17-1082} {{RACE:} large-scale reading comprehension dataset from examinations}.
\newblock In \emph{Proceedings of the 2017 Conference on Empirical Methods in Natural Language Processing, {EMNLP} 2017, Copenhagen, Denmark, September 9-11, 2017}, pages 785--794. Association for Computational Linguistics.

\bibitem[{Lee et~al.(2022)Lee, Ippolito, Nystrom, Zhang, Eck, Callison{-}Burch, and Carlini}]{DBLP:conf/acl/LeeINZECC22}
Katherine Lee, Daphne Ippolito, Andrew Nystrom, Chiyuan Zhang, Douglas Eck, Chris Callison{-}Burch, and Nicholas Carlini. 2022.
\newblock \href {https://doi.org/10.18653/V1/2022.ACL-LONG.577} {Deduplicating training data makes language models better}.
\newblock In \emph{Proceedings of the 60th Annual Meeting of the Association for Computational Linguistics (Volume 1: Long Papers), {ACL} 2022, Dublin, Ireland, May 22-27, 2022}, pages 8424--8445. Association for Computational Linguistics.

\bibitem[{Lefaudeux et~al.(2022)Lefaudeux, Massa, Liskovich, Xiong, Caggiano, Naren, Xu, Hu, Tintore, Zhang, Labatut, and Haziza}]{xFormers2022}
Benjamin Lefaudeux, Francisco Massa, Diana Liskovich, Wenhan Xiong, Vittorio Caggiano, Sean Naren, Min Xu, Jieru Hu, Marta Tintore, Susan Zhang, Patrick Labatut, and Daniel Haziza. 2022.
\newblock xformers: A modular and hackable transformer modelling library.
\newblock \url{https://github.com/facebookresearch/xformers}.

\bibitem[{Lehmann et~al.(2015)Lehmann, Isele, Jakob, Jentzsch, Kontokostas, Mendes, Hellmann, Morsey, van Kleef, Auer, and Bizer}]{dbpedia}
Jens Lehmann, Robert Isele, Max Jakob, Anja Jentzsch, Dimitris Kontokostas, Pablo~N. Mendes, Sebastian Hellmann, Mohamed Morsey, Patrick van Kleef, S{\"{o}}ren Auer, and Christian Bizer. 2015.
\newblock \href {https://doi.org/10.3233/SW-140134} {Dbpedia - {A} large-scale, multilingual knowledge base extracted from wikipedia}.
\newblock \emph{Semantic Web}, 6(2):167--195.

\bibitem[{Levine et~al.(2022)Levine, Wies, Jannai, Navon, Hoshen, and Shashua}]{pretraining-example-design}
Yoav Levine, Noam Wies, Daniel Jannai, Dan Navon, Yedid Hoshen, and Amnon Shashua. 2022.
\newblock \href {https://openreview.net/forum?id=lnEaqbTJIRz} {The inductive bias of in-context learning: Rethinking pretraining example design}.
\newblock In \emph{The Tenth International Conference on Learning Representations, {ICLR} 2022, Virtual Event, April 25-29, 2022}. OpenReview.net.

\bibitem[{Liu et~al.(2023{\natexlab{a}})Liu, Lin, Hewitt, Paranjape, Bevilacqua, Petroni, and Liang}]{liu2023lost}
Nelson~F. Liu, Kevin Lin, John Hewitt, Ashwin Paranjape, Michele Bevilacqua, Fabio Petroni, and Percy Liang. 2023{\natexlab{a}}.
\newblock \href {http://arxiv.org/abs/2307.03172} {Lost in the middle: How language models use long contexts}.

\bibitem[{Liu et~al.(2023{\natexlab{b}})Liu, Qiao, Neiswanger, Wang, Tan, Tao, Li, Wang, Sun, Pangarkar et~al.}]{llm360}
Zhengzhong Liu, Aurick Qiao, Willie Neiswanger, Hongyi Wang, Bowen Tan, Tianhua Tao, Junbo Li, Yuqi Wang, Suqi Sun, Omkar Pangarkar, et~al. 2023{\natexlab{b}}.
\newblock Llm360: Towards fully transparent open-source llms.
\newblock \emph{arXiv preprint arXiv:2312.06550}.

\bibitem[{Ott et~al.(2019)Ott, Edunov, Baevski, Fan, Gross, Ng, Grangier, and Auli}]{fairseq}
Myle Ott, Sergey Edunov, Alexei Baevski, Angela Fan, Sam Gross, Nathan Ng, David Grangier, and Michael Auli. 2019.
\newblock fairseq: A fast, extensible toolkit for sequence modeling.
\newblock In \emph{Proceedings of NAACL-HLT 2019: Demonstrations}.

\bibitem[{Pagliardini et~al.(2023)Pagliardini, Paliotta, Jaggi, and Fleuret}]{sparse-flash-attention}
Matteo Pagliardini, Daniele Paliotta, Martin Jaggi, and Fran{\c{c}}ois Fleuret. 2023.
\newblock \href {https://doi.org/10.48550/ARXIV.2306.01160} {Faster causal attention over large sequences through sparse flash attention}.
\newblock \emph{CoRR}, abs/2306.01160.

\bibitem[{Rajpurkar et~al.(2016)Rajpurkar, Zhang, Lopyrev, and Liang}]{squad}
Pranav Rajpurkar, Jian Zhang, Konstantin Lopyrev, and Percy Liang. 2016.
\newblock \href {https://doi.org/10.18653/V1/D16-1264} {Squad: 100, 000+ questions for machine comprehension of text}.
\newblock In \emph{Proceedings of the 2016 Conference on Empirical Methods in Natural Language Processing, {EMNLP} 2016, Austin, Texas, USA, November 1-4, 2016}, pages 2383--2392. The Association for Computational Linguistics.

\bibitem[{Shi et~al.(2023)Shi, Min, Lomeli, Zhou, Li, Lin, Smith, Zettlemoyer, Yih, and Lewis}]{in-context-pretraining}
Weijia Shi, Sewon Min, Maria Lomeli, Chunting Zhou, Margaret Li, Victoria Lin, Noah~A Smith, Luke Zettlemoyer, Scott Yih, and Mike Lewis. 2023.
\newblock In-context pretraining: Language modeling beyond document boundaries.
\newblock \emph{arXiv preprint arXiv:2310.10638}.

\bibitem[{Shin et~al.(2022)Shin, Lee, Ahn, Kim, Kim, Kim, Cho, Lee, Park, Ha, and Sung}]{DBLP:conf/naacl/ShinLAKKKCLPHS22}
Seongjin Shin, Sang{-}Woo Lee, Hwijeen Ahn, Sungdong Kim, HyoungSeok Kim, Boseop Kim, Kyunghyun Cho, Gichang Lee, Woo{-}Myoung Park, Jung{-}Woo Ha, and Nako Sung. 2022.
\newblock \href {https://doi.org/10.18653/V1/2022.NAACL-MAIN.380} {On the effect of pretraining corpora on in-context learning by a large-scale language model}.
\newblock In \emph{Proceedings of the 2022 Conference of the North American Chapter of the Association for Computational Linguistics: Human Language Technologies, {NAACL} 2022, Seattle, WA, United States, July 10-15, 2022}, pages 5168--5186. Association for Computational Linguistics.

\bibitem[{Shoeybi et~al.(2019)Shoeybi, Patwary, Puri, LeGresley, Casper, and Catanzaro}]{megatron}
Mohammad Shoeybi, Mostofa Patwary, Raul Puri, Patrick LeGresley, Jared Casper, and Bryan Catanzaro. 2019.
\newblock \href {http://arxiv.org/abs/1909.08053} {Megatron-lm: Training multi-billion parameter language models using model parallelism}.
\newblock \emph{CoRR}, abs/1909.08053.

\bibitem[{Soboleva et~al.(2023)Soboleva, Al-Khateeb, Myers, Steeves, Hestness, and Dey}]{slimpajama}
Daria Soboleva, Faisal Al-Khateeb, Robert Myers, Jacob~R Steeves, Joel Hestness, and Nolan Dey. 2023.
\newblock \href {https://www.cerebras.net/blog/slimpajama-a-627b-token-cleaned-and-deduplicated-version-of-redpajama} {{SlimPajama: A 627B token cleaned and deduplicated version of RedPajama}}.

\bibitem[{Socher et~al.(2013)Socher, Perelygin, Wu, Chuang, Manning, Ng, and Potts}]{sst2}
Richard Socher, Alex Perelygin, Jean Wu, Jason Chuang, Christopher~D. Manning, Andrew~Y. Ng, and Christopher Potts. 2013.
\newblock \href {https://aclanthology.org/D13-1170/} {Recursive deep models for semantic compositionality over a sentiment treebank}.
\newblock In \emph{Proceedings of the 2013 Conference on Empirical Methods in Natural Language Processing, {EMNLP} 2013, 18-21 October 2013, Grand Hyatt Seattle, Seattle, Washington, USA, {A} meeting of SIGDAT, a Special Interest Group of the {ACL}}, pages 1631--1642. {ACL}.

\bibitem[{Tirumala et~al.(2023)Tirumala, Simig, Aghajanyan, and Morcos}]{d4}
Kushal Tirumala, Daniel Simig, Armen Aghajanyan, and Ari~S. Morcos. 2023.
\newblock \href {https://doi.org/10.48550/ARXIV.2308.12284} {{D4:} improving {LLM} pretraining via document de-duplication and diversification}.
\newblock \emph{CoRR}, abs/2308.12284.

\bibitem[{Touvron et~al.(2023)Touvron, Lavril, Izacard, Martinet, Lachaux, Lacroix, Rozi{\`{e}}re, Goyal, Hambro, Azhar, Rodriguez, Joulin, Grave, and Lample}]{llama}
Hugo Touvron, Thibaut Lavril, Gautier Izacard, Xavier Martinet, Marie{-}Anne Lachaux, Timoth{\'{e}}e Lacroix, Baptiste Rozi{\`{e}}re, Naman Goyal, Eric Hambro, Faisal Azhar, Aur{\'{e}}lien Rodriguez, Armand Joulin, Edouard Grave, and Guillaume Lample. 2023.
\newblock \href {https://doi.org/10.48550/ARXIV.2302.13971} {Llama: Open and efficient foundation language models}.
\newblock \emph{CoRR}, abs/2302.13971.

\bibitem[{Tworkowski et~al.(2023)Tworkowski, Staniszewski, Pacek, Wu, Michalewski, and Milos}]{longllama}
Szymon Tworkowski, Konrad Staniszewski, Mikolaj Pacek, Yuhuai Wu, Henryk Michalewski, and Piotr Milos. 2023.
\newblock \href {https://doi.org/10.48550/ARXIV.2307.03170} {Focused transformer: Contrastive training for context scaling}.
\newblock \emph{CoRR}, abs/2307.03170.

\bibitem[{Xie et~al.(2023)Xie, Pham, Dong, Du, Liu, Lu, Liang, Le, Ma, and Yu}]{doremi}
Sang~Michael Xie, Hieu Pham, Xuanyi Dong, Nan Du, Hanxiao Liu, Yifeng Lu, Percy Liang, Quoc~V. Le, Tengyu Ma, and Adams~Wei Yu. 2023.
\newblock \href {https://doi.org/10.48550/ARXIV.2305.10429} {Doremi: Optimizing data mixtures speeds up language model pretraining}.
\newblock \emph{CoRR}, abs/2305.10429.

\bibitem[{Yang et~al.(2018)Yang, Qi, Zhang, Bengio, Cohen, Salakhutdinov, and Manning}]{hotpotqa}
Zhilin Yang, Peng Qi, Saizheng Zhang, Yoshua Bengio, William~W. Cohen, Ruslan Salakhutdinov, and Christopher~D. Manning. 2018.
\newblock \href {https://doi.org/10.18653/V1/D18-1259} {Hotpotqa: {A} dataset for diverse, explainable multi-hop question answering}.
\newblock In \emph{Proceedings of the 2018 Conference on Empirical Methods in Natural Language Processing, Brussels, Belgium, October 31 - November 4, 2018}, pages 2369--2380. Association for Computational Linguistics.

\bibitem[{Zhang et~al.(2024)Zhang, Zeng, Wang, and Lu}]{tinyllama}
Peiyuan Zhang, Guangtao Zeng, Tianduo Wang, and Wei Lu. 2024.
\newblock Tinyllama: An open-source small language model.
\newblock \emph{arXiv preprint arXiv:2401.02385}.

\bibitem[{Zhang et~al.(2022)Zhang, Roller, Goyal, Artetxe, Chen, Chen, Dewan, Diab, Li, Lin, Mihaylov, Ott, Shleifer, Shuster, Simig, Koura, Sridhar, Wang, and Zettlemoyer}]{opt}
Susan Zhang, Stephen Roller, Naman Goyal, Mikel Artetxe, Moya Chen, Shuohui Chen, Christopher Dewan, Mona~T. Diab, Xian Li, Xi~Victoria Lin, Todor Mihaylov, Myle Ott, Sam Shleifer, Kurt Shuster, Daniel Simig, Punit~Singh Koura, Anjali Sridhar, Tianlu Wang, and Luke Zettlemoyer. 2022.
\newblock \href {https://doi.org/10.48550/ARXIV.2205.01068} {{OPT:} open pre-trained transformer language models}.
\newblock \emph{CoRR}, abs/2205.01068.

\bibitem[{Zhang et~al.(2015)Zhang, Zhao, and LeCun}]{yelp_amazon_agnews}
Xiang Zhang, Junbo~Jake Zhao, and Yann LeCun. 2015.
\newblock \href {https://proceedings.neurips.cc/paper/2015/hash/250cf8b51c773f3f8dc8b4be867a9a02-Abstract.html} {Character-level convolutional networks for text classification}.
\newblock In \emph{Advances in Neural Information Processing Systems 28: Annual Conference on Neural Information Processing Systems 2015, December 7-12, 2015, Montreal, Quebec, Canada}, pages 649--657.

\end{thebibliography}

\clearpage

\appendix

\label{sec:appendix}

\clearpage

\section{Implementation of Intra-Document Masking}
\label{sec:speed}

We use FlashAttention2~\citep{flash-attention} to implement \intraMask. The pseudo-code is presented as follows:
\begin{algorithm}[H]
\label{alg:masking}
\caption*{~\footnotesize Pseudo-code for intra-document causal masking}
\begin{lstlisting}[language=python]
  # qkv_states: query, key and value
  # max_seqlen: max length of documents
  # cu_seqlens: boundaries of documents

  qkv_states = qkv_project(hidden_states)
  
  qkv_states = qkv_states.view(batch_size, seq_len, 3, num_heads, head_dim)
  
  qkv_states = rotary_embed(qkv_states)
  
  qkv_states = qkv_states.view(batch_size * seq_len, 3, num_heads, head_dim)
  
  attn = flash_attn_var_len_qkvpacked_func(qkv_states, cu_seqlens, max_seqlen, causal=True)
  
  attn = attn.view(batch_size, seq_len, num_heads * head_dim)
  attn = output_project(attn)
  
\end{lstlisting}
\end{algorithm}
In this implementation of \intraMask, we first apply the rotary position embedding to the hidden states, ensuring \UniChunkMask uses the same position information that is used in causal masking for each document.

We observe a $4\%$ pre-training speed decrease in our implementation compared to causal masking pre-training, testing on 128 80G A100 GPUs.
Another choice to implement \intraMask is using a binary attention bias matrix for masking tokens that belong to other documents. Compared to causal masking using FlashAttention2, we observe that it reduces efficiency by $32\%$ in xFormers~\citep{xFormers2022} when applying the attention bias; besides, it reduces efficiency by $52\%$ using the standard PyTorch implementation.

\section{Pre-Training Details}
\label{sec:pre-training detail}

\paragraph{Hyperparameters}
In our experiments, we use the $1.3$B model, which has $24$ layers, a hidden size of $2048$, and $16$ attention heads.
We use a batch size of $4$ million tokens for both the models with 2K and 8K context window sizes and pre-train models using $150$B tokens with $38400$ steps, which costs $40$ hours to pre-training a causal masking model using 128 80G A100 GPUs.
We use Adam optimiser with $\beta_1=0.90$, $\beta_2=0.95$, a weight decay of $0.1$, and a cosine learning rate scheduler.
The peak learning rate is $3\times 10^{-4}$, decreasing to $3\times 10^{-5}$ at the end.

\paragraph{Pre-Training Corpus}
\label{sec:pre-training data details}

We sample documents with $150$B tokens sampled from SlimPajama for pre-training.
All models are pre-trained using the same set of documents.
In \cref{tab:slimpajama-data-dtail}, we present the number of documents and the token proportions for each subset.

\begin{table}[t]
\centering
\small
\begin{tabular}{ccc} 
\toprule
\bf Subset & \bf \verb|#| documents  & \bf Token proportion\\
\midrule
CommonCrawl & $42960927$ & $52.2\%$ \\ 
C4 & $76520211$ & $26.7\%$ \\
GitHub & 5233374 & $5.2\%$ \\
Books & 47848 & $4.2\%$ \\
ArXiv & 383058 & $4.6\%$ \\
Wikipedia & 7044397 & $3.8\%$ \\
StackExchange & 7265708 & $3.3\%$ \\
\bottomrule
\end{tabular}
\caption{Pre-training corpus.}
\label{tab:slimpajama-data-dtail}
\end{table}

\section{Analysis of BM25Chunk}
\label{sec:bm25-detail}
\subsection{Time Complexity Analysis}
In BM25, the similarity score between a query and a document is based on sparse representations, where each query and document is represented by the terms it contains; such sparse representations are stored in \emph{inverted indices}, which map terms to the documents that contain them, along with necessary statistics such as the term frequency and the document frequency.
The time complexity of computing similarities between a query and documents in BM25 using an inverted index is $\mathcal{O}(Q \times K)$, where $Q$ denotes the number of tokens in the query, and $K$ represents the number of total documents.

To improve efficiency, we restrict \BMChunk's retrieval process within a document buffer rather than entire large-scale corpora.
The buffer caches $k$ documents, which enables similarity calculations between a term and documents to be at most $k$ times.
Since each query is a document, it could contain a large number of tokens; we remove the stop words and randomly sample $q$ tokens to reduce the length.
Therefore, the time complexity of sequence construction in \BMChunk is reduced to $O(q \times k)$. 
In~\cref{fig:construction_speed}, we test the sequence construction speed using different $q$ and $k$.

\subsection{Implementation Details}
\begin{figure}
    \centering
    \includegraphics[width=0.75\columnwidth]{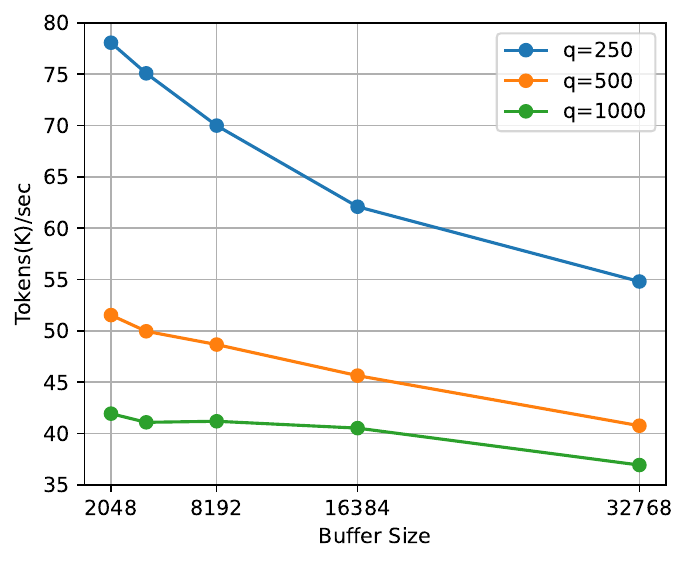}
    \caption{Pre-training sequence construction speeds using different buffer sizes $k$ and maximum query lengths $q$. Test on $16$ CPU cores.}
    \label{fig:construction_speed}
\end{figure}

We randomly group documents in batches of 5000K and build indexes within each group.
The BM25 indexes of pre-training corpora with $150$B tokens require $244$GB storage memory.
For both 2K and 8K settings, the document buffer size $k$ is $3072$, and the maximum length of query $q$ is $500$.
The data construction speed is $50.0$K tokens per second using $16$ CPU cores, and speeds using different settings are presented in \cref{fig:construction_speed}.

\subsection{Ablation Studies}
\paragraph{Effectiveness of Document Buffer}
\label{sec:bm25_analysis}
\BMChunk conducts retrieval within a document buffer, which may result in retrieving less relevant documents, so we conduct experiments on different document buffer sizes to investigate its effectiveness.
We conduct ablation experiments using $0.3$B models with a context window of $2048$, trained with 13B tokens, the compute-optimal number of tokens according to \citet{chinchilla}.
We present the PPL improvement over \UniChunk on the validation set of SlimPajama in~\cref{tab:document_buffer_size}.
The results show that retrieving from different sizes of document buffers can improve PPL, indicating the effectiveness of retrieving from a small-scale document set.
\BMChunk with a buffer size of $4096$ achieves the best result, while increasing the size to $8192$ does not improve the PPL.

\begin{table}[t]
\centering
\small
\resizebox{0.9\linewidth}{!}{
\begin{tabular}{lcl}
\toprule
{\bf Model} (0.3B) & \bf \makecell[c]{Document \\ Buffer Size} & \bf  Valid. PPL \\ \midrule
\MixChunk & - & $15.474$\\ 
\UniChunkMask & - & $12.443_{\downarrow 3.031}$ \\
\midrule
\multirow[m]{3}{*}{\BMChunk} & 2048 & $13.657_{\downarrow 1.817}$ \\
& 4096 & $\textbf{12.528}_{\downarrow 2.946}$ \\
&  8192 & $12.684_{\downarrow 2.790}$ \\ 
\midrule
\BMChunk \\
\makecell[l]{\quad w/o multi-hop retrieval} & 4096 & $13.497_{\downarrow 1.977}$ \\
\makecell[l]{\quad w/o retrieval} & 4096 & $14.241_{\downarrow 1.233}$ \\ 
\midrule
\textsc{Contriever}Chunk & - & $13.720_{\downarrow 1.654}$ \\
\bottomrule
\end{tabular}
}
\caption{PPL on the validation set of SlimPajama. Subscript$_\downarrow$ is the PPL improvement over \MixChunk. The label ``w/o multi-hop retrieval'' means retrieving multiple documents at once to construct the sequence; ``w/o retrieval'' represents random sampling from document buffers, which is equivalent to \UniChunk.
}
\label{tab:document_buffer_size}
\end{table}

\paragraph{Effectiveness of Retrieval}
\BMChunk conducts multi-hop retrieval to retrieve a sequence of documents, which could potentially help models learn long-distance relationships across documents, and this benefit has been revealed by its high accuracy on HotpotQA, a multi-hop QA task, as shown in~\cref{sec:mrc_and_rag}.
An alternative choice is retrieving multiple documents at once to fill a pre-training chunk, and we present such one-hop retrieval in \cref{tab:document_buffer_size}.
The result indicates that \BMChunk with multi-hop retrieval can improve the PPL more effectively. 
Besides, we experiment with random sampling documents from the buffers without retrieval; the result shows the effectiveness of retrieval.

\paragraph{Dense Retrieval Method}

An alternative retrieval method to BM25 is dense retrieval.
We use Contreiver~\citep{contriever} as the dense retriever and compare it with BM25.
Following~\citet{in-context-pretraining}, we embed pre-training documents to dense vectors using Contriever and use FAISS~\citep{faiss} to accelerate the retrieval process instead of using the document buffer.
Then, we construct pre-training chunks using the same process introduced in \BMChunk.
We present the result produced by the dense retrieval method in the last line of \cref{tab:document_buffer_size}.
We observe that the improvement of the dense retrieval method is less than BM25.

\section{Analysis of Data Distribution Properties}
\label{sec:appendix-distribution-property}

\begin{figure}
    \centering
    \includegraphics[width=0.8\columnwidth]{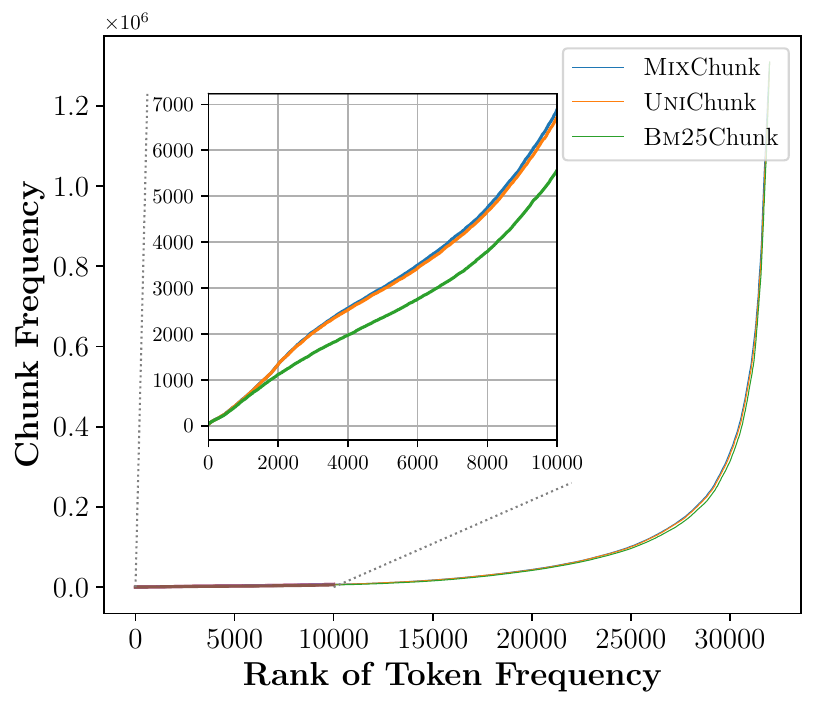}
    \caption{
Chunk frequency. The $x$-axis indicates the frequency rank of tokens; the $y$-axis presents the number of chunks containing a specific token.
}
    \label{fig:chunk_frequency}
\end{figure}

\paragraph{Chunk Frequency}
In addition to Zipf's coefficient, we analyse the burstiness property through the chunk frequency of tokens.
Specifically, chunk frequency refers to the number of chunks where a specific token appears. 
Given a corpus, if a specific token appears in fewer chunks, it indicates more concentrated occurrences in chunks containing the token, demonstrating a higher burstiness property.
In \cref{fig:chunk_frequency}, we can see that low-frequency tokens appear in fewer chunks in \BMChunk compared to \MixChunk and \UniChunk, indicating these low-frequency tokens are gathered through the retrieval-based construction method.

\paragraph{Distinct N-gram}
The burstiness property can correlate to the duplication in a sequence, which may negatively affect models, e.g., models may tend to copy phrases from context.
We use SlimPajama, a high-quality and deduplicated dataset, as the pre-training corpus, which can alleviate the duplication issue in \BMChunk.
We use the percentage of distinct n-grams within a sequence to analyse the duplication issue, as shown in \cref{tab:distinct-n-gram}.
The results show that, with \BMChunk, pre-training sequences contain a lower percentage of distinct n-grams than \MixChunk and \UniChunk.

\begin{figure*}[ht]
\small
    \centering
    \begin{subfigure}[b]{0.24\textwidth}
    \includegraphics[width=1\linewidth]{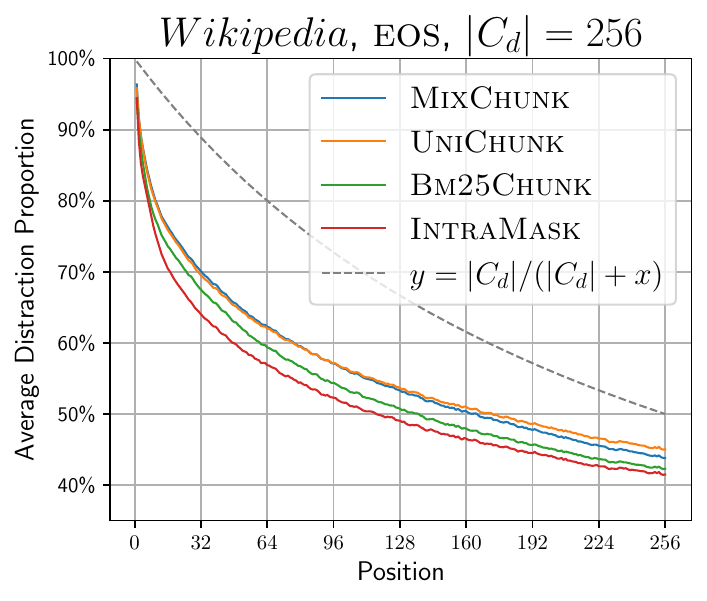}
    \caption{}
    \end{subfigure}
        \begin{subfigure}[b]{0.24\textwidth}
    \includegraphics[width=1\linewidth]{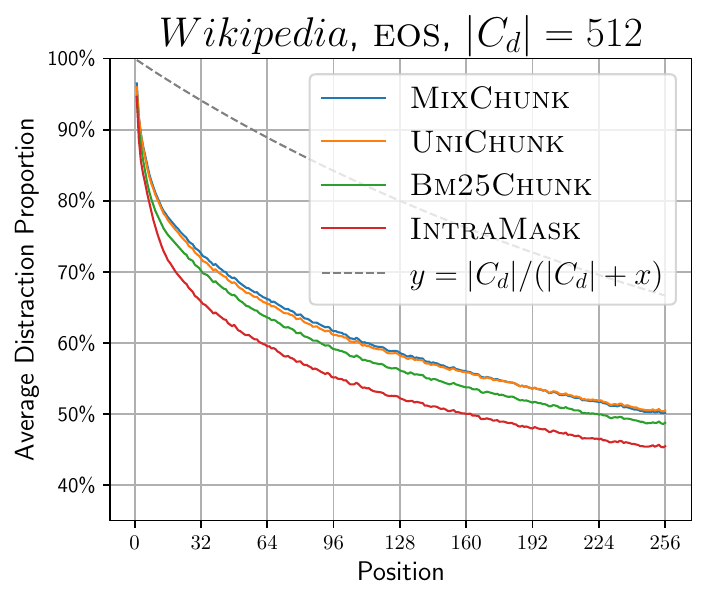}
    \caption{}
    \end{subfigure}
        \begin{subfigure}[b]{0.24\textwidth}
    \includegraphics[width=1\linewidth]{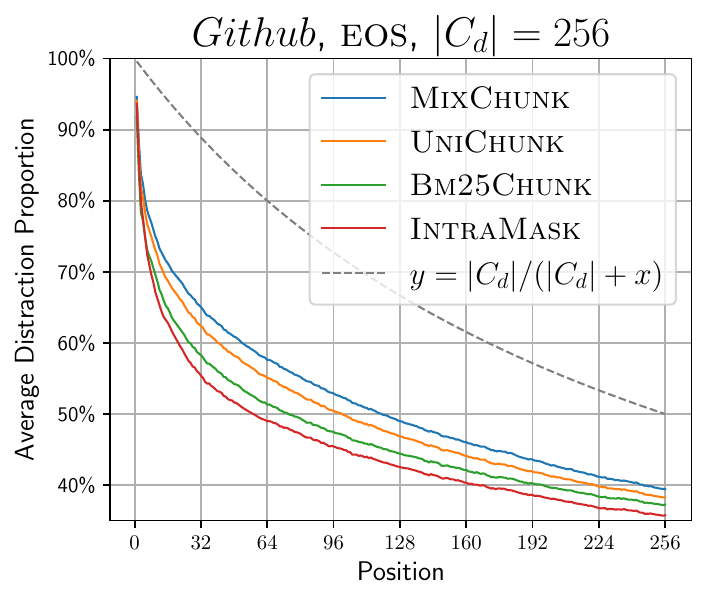}  
    \caption{}
    \end{subfigure}
        \begin{subfigure}[b]{0.24\textwidth}
    \includegraphics[width=1\linewidth]{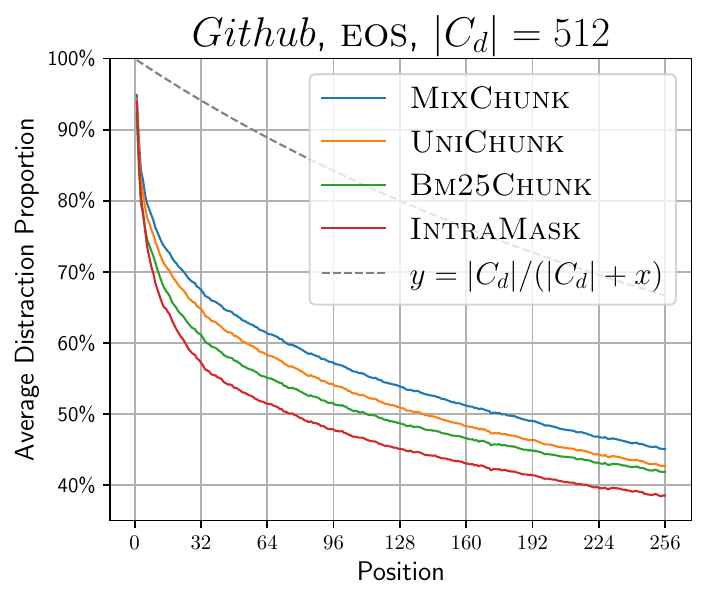}
    \caption{}
    \end{subfigure}
\\
    \begin{subfigure}[b]{0.24\textwidth}
    \includegraphics[width=1\linewidth]{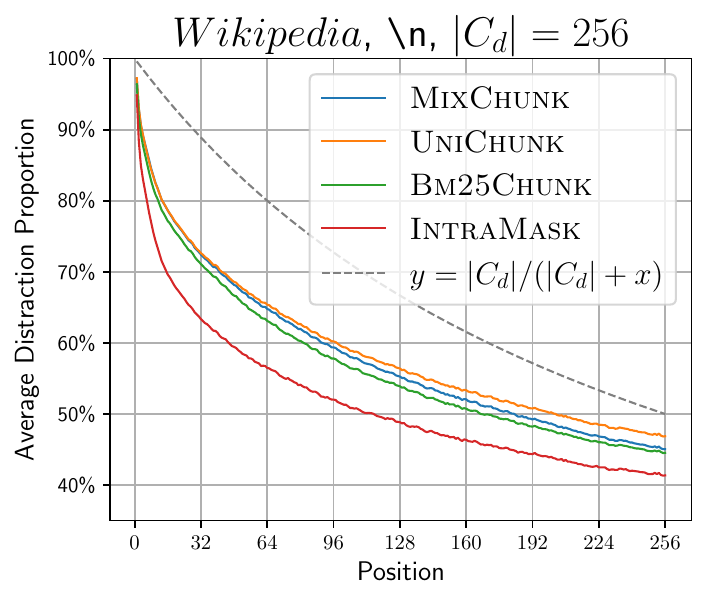}
    \caption{}
    \end{subfigure}
        \begin{subfigure}[b]{0.24\textwidth}
    \includegraphics[width=1\linewidth]{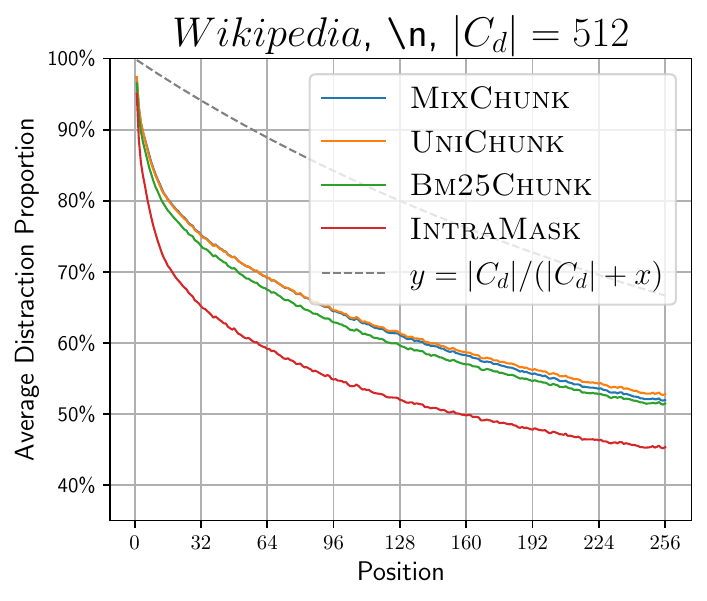}
    \caption{}
    \end{subfigure}
        \begin{subfigure}[b]{0.24\textwidth}
    \includegraphics[width=1\linewidth]{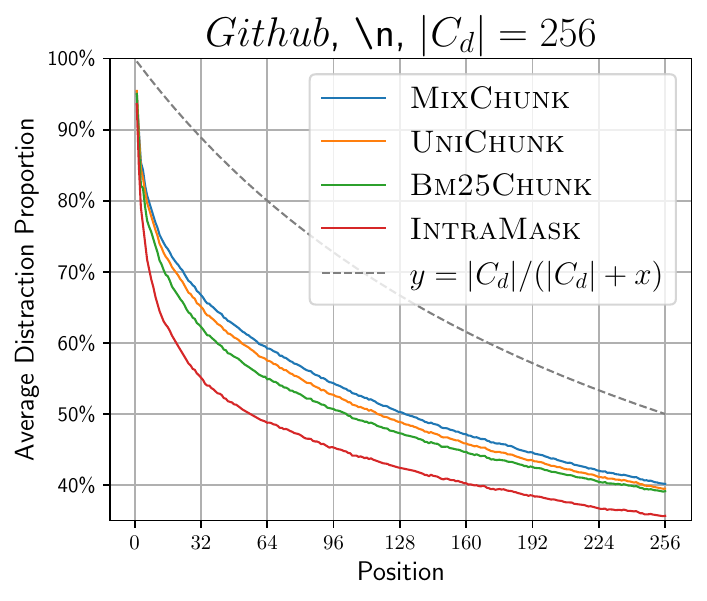}
    \caption{}
    \end{subfigure}
        \begin{subfigure}[b]{0.24\textwidth}
    \includegraphics[width=1\linewidth]{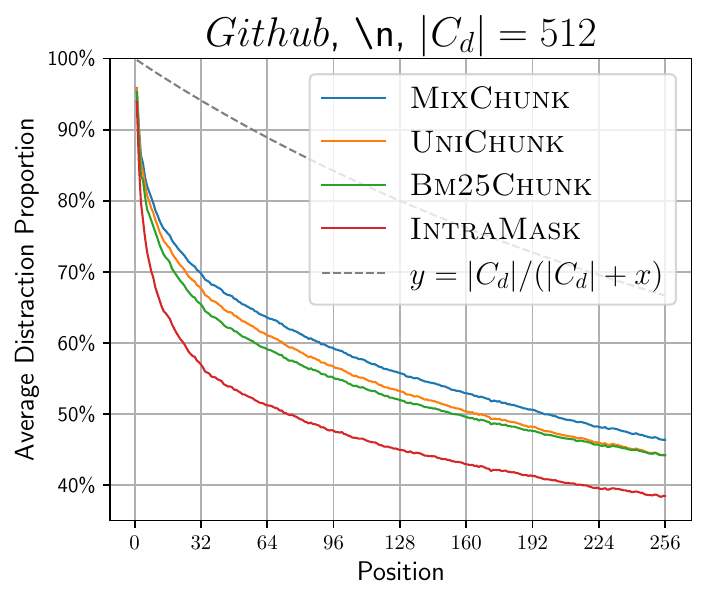}
    \caption{}
    \end{subfigure}

\caption{
Average distraction proportions over layers. We compare results using different corpora (Wikipedia and GitHub), distraction length ($|C_d|=256$ and $512$), and the separator \textsc{[eos]} and $\backslash \text{n}$).
The first row, (a) (b) (c) and (d), use \textsc{[eos]} as the separator; the second row, (e) (f) (g) and (h), use $\backslash \text{n}$. The first and the third columns, (a) (c) (e) and (g), have an irrelevant context length $|C_d|$ of $256$, and the others are $512$. The first two columns, (a) (b) (e) and (f), present the results of Wikipedia, and the last two columns, (c) (d) (g) and (h), present the results of GitHub.
We present the baseline $y=|C_d|/(|C_d|+x)$ whose attention scores are uniformly distributed over all preceding tokens.
}
    \label{fig:more_distraction_results}
\end{figure*}

\begin{table}[t]
\centering
\small
\setlength\tabcolsep{10pt}
\begin{tabular}{lcc}
\toprule
Method & $\Delta$ PPL $\%$ & $\Delta$ \textsc{DistProp} $\%$ \\  \midrule
\MixChunk & $14.6\%$ & $3.4\%$ \\
\UniChunk & $15.3\%$ & $4.6\%$ \\
\BMChunk  & $13.5\%$ & $4.6\%$ \\
\UniChunkMask & $-0.7\%$ & $-0.6\%$ \\
\bottomrule
\end{tabular}
\caption{The PPL and \textsc{DistProp} changes after replacing the separator \textsc{[eos]} by $\backslash \text{n}$. A positive value means PPL or \textsc{DistProp} increases (performance drops).}
\label{tab:distraction_delta}
\end{table}
\section{Analysis of Distraction Proportions in Different Settings}
\label{sec:more_distraction}
In \cref{fig:more_distraction_results}, we report the average distraction proportion (defined in \cref{eq:distraction}) over layers using different settings.
Specifically, we analyse distraction proportions in different settings by varying the \textit{1)} modalities of corpus: text and code using Wikipedia and GitHub; \textit{2)} the separator token: \textsc{[eos]} and line break token $\backslash \text{n}$; \textit{3)} the length of distraction context, $|C_d|=256$ and $512$.

Comparing different separators \textsc{[eos]} and $\backslash \text{n}$, (a) (e), (b) (f), (c) (g), and (d) (h), we observe that causal masking models can obtain lower distraction proportions using \textsc{[eos]}, indicating causal masking models can benefit from \textsc{[eos]} to ignore irrelevant context during pre-training. 
We present the impact of changing the separator from \textsc{[eos]} to $\backslash \text{n}$ on PPL and distraction proportion in \cref{tab:distraction_delta}.
The results show that PPL and \textsc{DistProp} increase after the replacement for causal masking models, while \UniChunkMask obtains better results using $\backslash \text{n}$ as the separator since it does not train on sequences where documents are separated by \textsc{[eos]} using \intraMask.

Comparing Wikipedia (a) (b) (e) (f) and GitHub (c) (d) (g) (h), \MixChunk is more distracted by the irrelevant context in code generation.

Comparing different length distraction contexts, (a) (b), (c) (d), (e) (f) and (g) (h), models are more distracted when $|C_d|$ increases, while much better than the baseline of uniform distribution $y=|C_d|/(|C_d|+x)$.

Comparing \UniChunkMask (red line) and causal masking models, we observe that \intraMask results in significantly lower distraction proportions in all cases.
This phenomenon may imply that using causal masking without considering the boundaries of documents negatively impacts language modelling performance, and the models can be more robust to irrelevant contexts when increasing the relatedness of documents in pre-training chunks.

\section{Next Token Accuracy of Pre-Trained Language Models}
\label{sec:next-token-accuracy}

\begin{table*}[htbp]
\centering
\resizebox{\textwidth}{!}{
\begin{tabular}{clccccccc|c}
\toprule
$L$ & \bf Model & \bf CommonCrawl & \bf C4 & \bf Wikipedia & \bf GitHub & \bf StackExchange & \bf Book & \bf ArXiv & \bf Avg. \\ \midrule
\multirow[m]{4}{*}{2K} 
& \MixChunk & $0.5429$ & $0.4950$ & $0.6238$ & $0.7665$ & $0.5974$ & $0.5001$ & $0.6406$ & $0.5952$ \\ 
& \UniChunk & $0.5468$ & $0.4984$ & $0.6298$ & $0.7709$ & $0.6011$ & $0.5033$ & $0.6436$ & $0.5991$ \\
& \BMChunk & $\underline{0.5496}$ & $\underline{0.5021}$ & $\underline{0.6394}$ & $\underline{0.7782}$ & $\underline{0.6041}$ & $\underline{0.5050}$ & $\underline{0.6452}$ & $\underline{0.6034}$  \\ 
& \UniChunkMask & $\textbf{0.5507}$ & $\textbf{0.5048}$ & $\textbf{0.6426}$ & $\textbf{0.7793}$ & $\textbf{0.6050}$ & $\textbf{0.5062}$ & $\textbf{0.6458}$ & $\textbf{0.6049}$ \\ 

\midrule

\multirow[m]{4}{*}{8K} & \MixChunk & $0.5402$ & $0.4867$ & $0.6219$ & $0.7443$ & $0.5820$ & $0.5042$ & $0.6531$ & $0.5903$ \\ 
& \UniChunk & $0.5429$ & $0.4888$ & $0.6235$ & $0.7483$ & $0.5859$ & $0.5065$ & $0.6564$ & $0.5932$ \\ 
& \BMChunk & $\underline{0.5489}$ & $\underline{0.4952}$ & $\underline{0.6391}$ & $\underline{0.7621}$ & $\underline{0.5919}$ & $\underline{0.5108}$ & $\textbf{0.6599}$ & $\underline{0.6011}$ \\ 
& \UniChunkMask & $\textbf{0.5506}$ & $\textbf{0.4988}$ & $\textbf{0.6443}$ & $\textbf{0.7643}$ & $\textbf{0.5936}$ & $\textbf{0.5119}$ & $\underline{0.6597}$ & $\textbf{0.6033}$ \\ 
\bottomrule
\end{tabular}
}
\caption{Evaluation of next token accuracy on SlimPajama's test set.
}
\label{tab:next-token-accuracy}
\end{table*}
In addition to PPL, we report the next token accuracy of pre-trained language models in~\cref{tab:next-token-accuracy}.

\begin{table}[t]
\centering
\small
\resizebox{\linewidth}{!}{
\begin{tabular}{clccc}
\toprule
 $L$ & \bf Method & \makecell{\bf Distinct \\ \bf 2-gram $\%$} & \makecell{\bf Distinct \\ \bf 3-gram $\%$} & \makecell{\bf Distinct \\ \bf 4-gram $\%$} \\ \midrule
\multirow[m]{4}{*}{2K} & \MixChunk & $71.84_{\pm 14.68}$ & $84.06_{\pm 14.47}$ & $89.02_{\pm 13.16}$ \\
& \UniChunk & $71.84_{\pm 15.07}$ & $84.17_{\pm 14.74}$ & $89.16_{\pm 13.26}$ \\
& \BMChunk & $71.49_{\pm 15.21}$ & $84.00_{\pm 14.91}$ & $89.07_{\pm 13.41}$ \\
& \UniChunkMask & $80.35_{\pm 15.26}$ & $89.01_{\pm 13.07}$ & $92.61_{\pm 11.34}$ \\ 
\midrule
\multirow[m]{4}{*}{8K} & \MixChunk & $64.81_{\pm 12.84}$ & $80.61_{\pm 13.69}$ & $86.76_{\pm 12.76}$ \\
& \UniChunk & $64.57_{\pm 14.09}$ & $80.61_{\pm 14.92}$ & $86.88_{\pm 13.64}$ \\
& \BMChunk & $63.49_{\pm 14.63}$ & $80.06_{\pm 15.64}$ & $86.56_{\pm 14.31}$ \\
& \UniChunkMask & $79.88_{\pm 14.86}$ & $88.90_{\pm 12.63}$ & $92.61_{\pm 10.96}$ \\ 
\bottomrule
\end{tabular}
}
\caption{The percentages of the distinct n-grams in different pre-training sequences.}
\label{tab:distinct-n-gram}
\end{table}

\end{document}